%% file: main_final.tex
\documentclass[lettersize,journal]{IEEEtran}
\usepackage{amsmath,amsfonts}
\usepackage{algorithmic}
\usepackage{algorithm}
\usepackage{listings}
\usepackage{float}
\usepackage{array}
\usepackage[caption=false,font=normalsize,labelfont=sf,textfont=sf]{subfig}
\usepackage{textcomp}
\usepackage{stfloats}
\usepackage{url}
\usepackage{verbatim}
\usepackage{graphicx}
\usepackage{cite}
\usepackage{booktabs}
\usepackage{multirow}
\usepackage{microtype}
\usepackage{needspace}
\usepackage{etoolbox}
\usepackage[table]{xcolor}
\usepackage{pifont}
\usepackage[colorlinks,linkcolor=blue,citecolor=blue,urlcolor=blue]{hyperref}
\definecolor{arxivcodeblue}{rgb}{0.07843,0.47843,1}
\usepackage[capitalise,noabbrev]{cleveref}

\graphicspath{{images/}}
\definecolor{updategreen}{RGB}{0,128,0}
\definecolor{algpink}{RGB}{255,20,147}
\definecolor{key_words}{HTML}{ec008d}
\definecolor{annotation}{HTML}{009900}
\definecolor{lightpurple}{RGB}{236,231,248}
\definecolor{tablepurple}{RGB}{234,228,248}
\definecolor{markred}{RGB}{220,0,0}
\definecolor{markgreen}{RGB}{0,175,0}

\newcommand{\cmark}{\textcolor{red}{\ding{51}}}
\newcommand{\xmark}{\textcolor{updategreen}{\ding{55}}}

\makeatletter
\def\ps@arxivplain{%
  \def\@oddhead{}%
  \let\@evenhead\@oddhead
  \def\@oddfoot{}\let\@evenfoot\@oddfoot}
\let\ps@headings\ps@arxivplain
\let\ps@titlepagestyle\ps@arxivplain
\let\ps@IEEEtitlepagestyle\ps@arxivplain
\makeatother
\AtBeginDocument{\pagestyle{arxivplain}}
\AddToHook{shipout/foreground}{\put(576,-34){\makebox(0,0)[r]{\normalfont\fontsize{8pt}{8pt}\selectfont\thepage}}}

\begin{document}

\title{SGLDet: Spectral-aware Global-Local Dynamic Modeling for Multispectral Aerial Object Detection}

\author{Peng Zhang,
        Shuaihao Han,
        Tingfa Xu\textsuperscript{*},
        and Jianan Li\textsuperscript{*}%
\thanks{Peng Zhang and Shuaihao Han are with Beijing Institute of Technology, Beijing 100081, China (e-mail: 3120240593@bit.edu.cn; 3120230561@bit.edu.cn).\protect\\\hspace*{1em}Tingfa Xu and Jianan Li are with Beijing Institute of Technology, Beijing 100081, China, and also with Chongqing Innovation Center, Beijing Institute of Technology, Chongqing 401135, China (e-mail: ciom\_xtf1@bit.edu.cn; lijianan@bit.edu.cn).\protect\\\hspace*{1em}Peng Zhang and Shuaihao Han contributed equally to this work.\protect\\\hspace*{1em}\textsuperscript{*}Corresponding authors: Tingfa Xu and Jianan Li.}%
}

\markboth{}{}

\maketitle

\input{sec/0_abstract}

\vspace{-11.02956pt}

\input{sec/1_intro}

\input{sec/2_relatedwork}

\input{sec/3_method}

\input{sec/4_experiments}

\input{sec/5_conclusion}

\renewcommand{\IEEEbibitemsep}{-0.8pt}
\bibliographystyle{IEEEtran}
\bibliography{main}

\raggedbottom

\end{document}

%% file: sec/0_abstract.tex
\begin{abstract}
Multispectral aerial object detection aims to exploit spectral--spatial cues to address object detection challenges in complex aerial scenes. Existing methods mainly focus on spectral--spatial decoupling and object-aware spectral--spatial fusion to improve spectral utilization. However, two critical issues remain underexplored: 1) severe foreground--background imbalance caused by large scene coverage hinders reliable object recognition; 2) aerial objects exhibit substantial local variations in scale, orientation, and structure, which hinder accurate localization. To address these issues, we propose SGLDet, a spectral-aware global--local dynamic modeling framework for multispectral aerial object detection, which is motivated by the observation that object semantics are jointly determined by global spectral consistency and scene context, while precise localization requires dynamic feature modulation conditioned on the input spectral and spatial properties. To realize these properties, SGLDet consists of three coordinated components: a Spectral-driven Global Dynamic Modeling module that aggregates spectrally related features across the scene via similarity-guided sparse matching; a Multi-scale Global Context Integration module that constructs a compact scene memory for scene-guided contextual reasoning; and a Spectral-aware Input-dependent Calibration Aggregation module that performs expert-kernel aggregation with scene-guided spectral--spatial calibration to enable adaptive local feature modulation. Extensive experiments on MODA, HOD3K, and DroneVehicle demonstrate that SGLDet achieves superior performance with competitive computational cost.
Code is available at {\hypersetup{urlcolor=arxivcodeblue}\url{https://github.com/Riiluo/SGLDet}}.
\end{abstract}

%% file: sec/1_intro.tex
\section{Introduction}
\label{sec:intro}

\IEEEPARstart{A}{erial} object detection is a fundamental task in remote sensing, with applications in urban management, traffic surveillance, maritime monitoring, and disaster response. Most existing methods rely on RGB inputs, whose limited spectral representation often struggles under illumination changes, occlusions, and background interference, especially when objects resemble surrounding regions. Multispectral images (MSIs) provide richer spectral responses across multiple wavelengths and thus offer complementary cues beyond spatial appearance, making them a promising information source for aerial object detection. Existing methods for multispectral aerial object detection (MAOD) mainly either decouple spectral and spatial information for separate modeling or strengthen adaptive spectral-spatial fusion for target perception \cite{he2023object, han2025moda, xu2026dahnet}. However, because these methods mainly focus on local spectral-spatial interaction and emphasize feature enhancement within limited receptive fields, they remain inherently insufficient for resolving global semantic ambiguity in complex real-world aerial scenes.

\begin{figure}[!t]
    \centering
    \includegraphics[width=\linewidth]{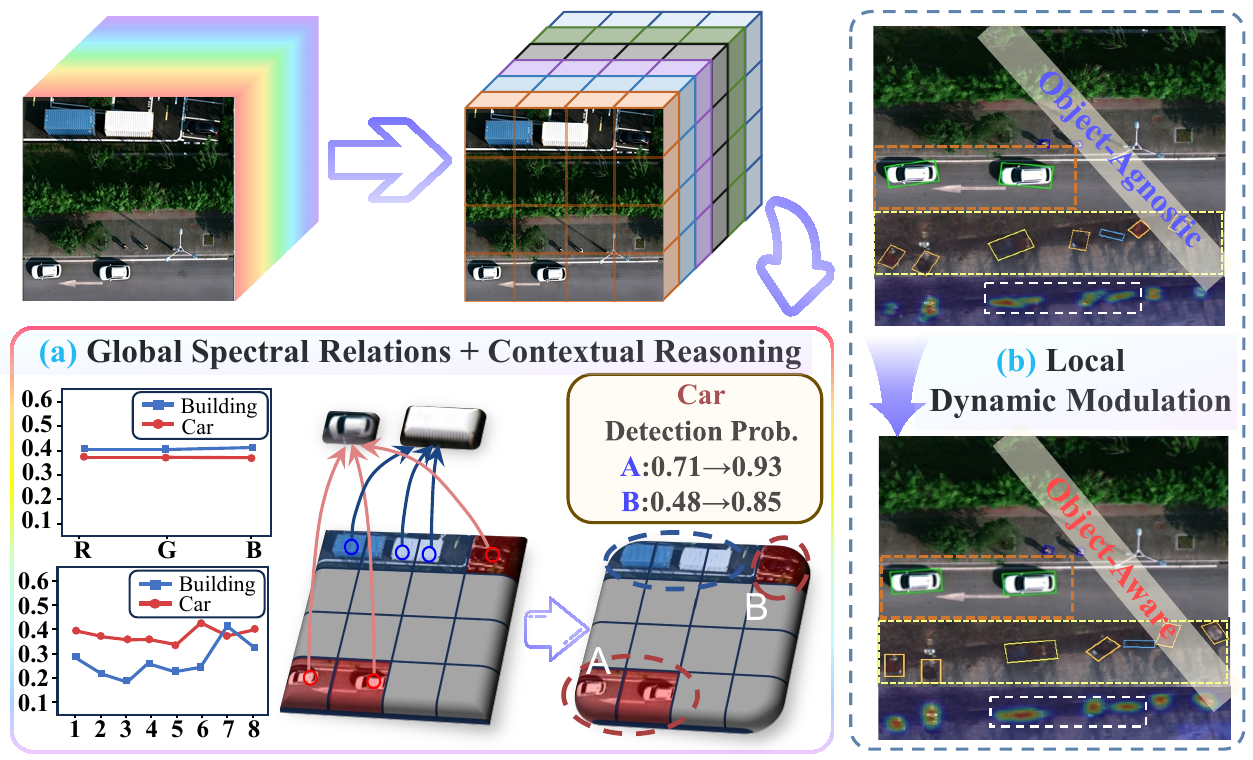}
    \caption{(a) Global spectral relations associate objects or regions with similar spectral responses across spatial locations, while global scene context further helps disambiguate visually confusing regions. SGDM and MGCI perform these two operations, respectively. (b) Local dynamic modulation produces more object-aligned responses and facilitates more accurate localization for objects in complex backgrounds. SICA implements this input-adaptive refinement.}
    \label{fig:fig1}
\end{figure}

Concretely, aerial images usually cover large spatial extents, and aerial objects often appear at small scales and are embedded in complex scenes. In MODA~\cite{han2025moda}, for example, objects occupy only 0.15\% of the image area on average, resulting in severe foreground--background imbalance, so local spectral--spatial cues alone are often insufficient for reliable recognition. Fortunately, MSI data provide a clue that is difficult to obtain from RGB imagery: objects composed of similar materials often exhibit consistent spectral signatures even across distant spatial locations, while scene context can further suppress visually confusing background regions. As shown in~\cref{fig:fig1}(a), linking distant regions with similar spectral responses can reinforce objects with ambiguous local appearances while suppressing visually similar but spectrally distinct background responses. Scene context further resolves the remaining semantic ambiguity. These observations indicate that robust MAOD should integrate scene-wide spectral relations with scene-level context to better resolve semantic ambiguity in aerial scenes. Motivated by this observation, we introduce two complementary designs for global reasoning in multispectral aerial object detection. The first is a Spectral-driven Global Dynamic Modeling (SGDM) module, which retrieves spectrally similar responses through similarity-guided sparse matching to encourage semantically coherent representations for spectrally similar objects. The second is a Multi-scale Global Context Integration (MGCI) module, which consolidates cross-scale evidence into a compact scene memory and enhances scene-guided contextual reasoning through contextual retrieval.

However, resolving global semantic ambiguity is still insufficient for MAOD. Aerial objects often vary substantially in scale, orientation, and structure, whereas standard convolutions apply fixed kernels and receptive fields to all inputs. This mismatch makes it difficult to produce object-aligned responses in cluttered backgrounds, as shown in~\cref{fig:fig1}(b). Local dynamic modulation therefore complements global reasoning by improving object localization. To address this issue, we further introduce a Spectral-aware Input-dependent Calibration Aggregation (SICA) module, which performs scene-dependent dynamic feature modulation through input-adaptive expert aggregation with spectral--spatial calibration for precise localization. Together, SGDM, MGCI, and SICA constitute SGLDet, a spectral-aware global-local dynamic modeling framework for multispectral aerial object detection.

Overall, SGLDet unifies global spectral dependency modeling, scene-guided contextual reasoning, and scene-dependent local feature modulation within a single framework. To evaluate the proposed method, we conduct extensive experiments on three benchmark datasets: MODA~\cite{han2025moda}, HOD3K~\cite{he2023object}, and DroneVehicle~\cite{dronevehicle}. The results demonstrate that SGLDet achieves superior detection performance with competitive computational cost. As shown in~\cref{fig:fig2}, SGLDet achieves stronger detection performance on both MODA and HOD3K with comparable computational cost. The main contributions are summarized as follows:

\begin{figure*}[!t]
    \centering
    \includegraphics[width=\textwidth]{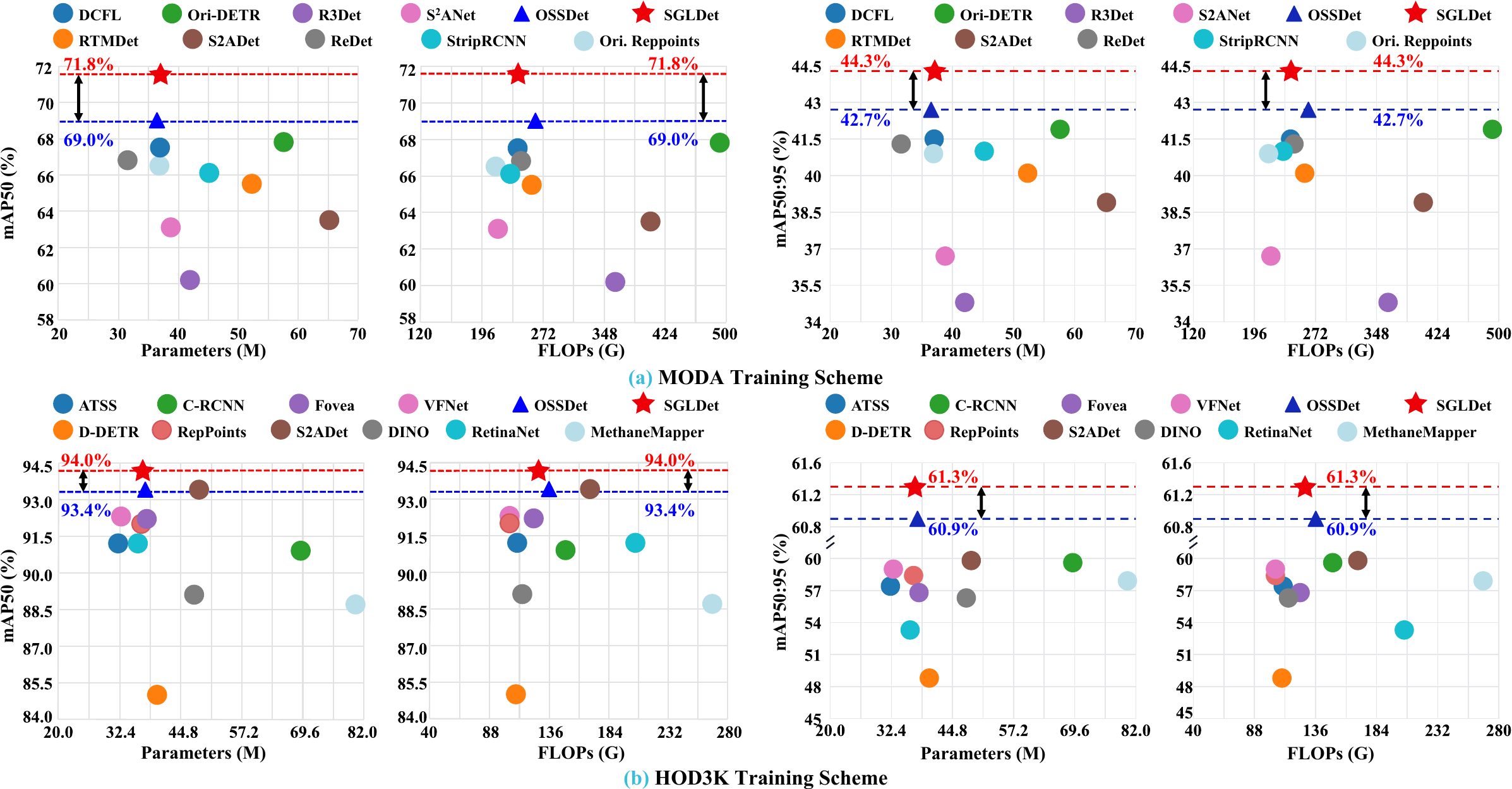}
    \caption{Accuracy--complexity comparisons on MODA and HOD3K using mAP$_{50}$ (left) and mAP (right).}
    \label{fig:fig2}
\end{figure*}
\begin{itemize}
    \item We propose SGLDet, a spectral-aware global-local dynamic modeling framework for multispectral aerial object detection that unifies global spectral dependency modeling, scene-guided contextual refinement, and scene-dependent dynamic local modulation.
    \item We develop a spectral-driven global reasoning design coupling similarity-guided sparse retrieval with scene-guided refinement, strengthening semantic consistency across spectrally similar objects and alleviating global semantic ambiguity in aerial scenes.
    \item We develop a spectral-aware input-dependent calibration aggregation mechanism with input-adaptive expert aggregation and spectral--spatial calibration for accurate localization under local variations.
\end{itemize}

%% file: sec/2_relatedwork.tex
\section{Related Work}
\label{sec:related_work}

\subsection{Object Detection in Aerial Images}
\label{sec:rw_aerial}

Object detection in RGB images has achieved remarkable progress in recent years, driven by advances in single-stage~\cite{tian2019fcos}, two-stage~\cite{Cascade}, and Transformer-based detectors~\cite{deformable-detr}. In aerial imagery, however, objects frequently appear with arbitrary orientations, necessitating specialized representations beyond horizontal bounding boxes. Two-stage approaches such as RoI Transformer~\cite{roi_trans} address rotation through region transformation, while single-stage oriented detectors such as R3Det~\cite{yang2021r3det}, S$^2$ANet~\cite{han2021align}, Oriented RepPoints~\cite{oriented-reppoints}, and CFA~\cite{cfa} improve oriented localization and dense object modeling. More recently, LSKNet~\cite{Li_2024_IJCV} and Strip R-CNN~\cite{yuan2025strip} explore large-kernel designs to better capture broader spatial context in remote sensing scenes.

Beyond geometric modeling, recent aerial detectors further improve dense-instance perception through stronger spatial context modeling and feature aggregation~\cite{Li_2024_IJCV, yuan2025strip, wu2025measuring}. However, these methods primarily rely on RGB spatial features and therefore remain limited when objects share similar visual appearances with cluttered backgrounds.

\subsection{Multispectral Object Detection}
\label{sec:rw_mod}

Leveraging information from continuous spectral bands, multispectral imagery holds significant potential for scene characterization and interpretation~\cite{hong2021spectralformer}. However, due to the limitations of multispectral image acquisition and resolution, datasets and methods for multispectral object detection remain relatively scarce. Multispectral data usually exhibit high correlation and redundancy among adjacent bands, and early studies mainly focused on pixel-level analysis such as spectral matching or anomaly detection, rather than object-level localization.

Yan et al.~\cite{yan2021object} explored spectral image object detection at the object level. He et al.~\cite{he2023object} proposed S2ADet and the HOD3K dataset, introducing a two-stream architecture that separately extracts spectral and spatial features before aggregating dual detection branches. Han et al.~\cite{han2025moda} further introduced MODA and proposed OSSDet to enhance target perception and suppress background interference through spectral-spatial interaction and object-aware cues. Xu et al.~\cite{xu2026dahnet} subsequently proposed DAHNet, which decomposes multispectral data and aggregates complementary spectral-spatial representations through cross-domain synergistic attention. Despite these advances, spectral information is still mainly used as an auxiliary cue for local feature enhancement through dimensionality reduction or channel reweighting, rather than as a global semantic prior for unifying materially consistent objects across the scene \cite{he2023object, han2025moda, xu2026dahnet}.

\subsection{Spatial-Spectral Feature Learning}
\label{sec:rw_ss}

Spatial-spectral feature learning has been extensively studied in multispectral image analysis, where the primary objective is to jointly model spectral correlations and spatial context for pixel-level discrimination. Early deep learning methods applied 2D CNNs to extract spatial features from individual bands or reduced representations, while 3D CNNs were later introduced to capture inter-band dependencies through volumetric convolutions. Despite these advances, most spatial-spectral learning paradigms are designed for classification or restoration rather than accurate instance-level localization in aerial detection.

When extended to object detection, spatial-spectral learning remains relatively underexplored. Existing methods usually treat spectral cues as auxiliary signals in feature fusion \cite{he2023object, han2025moda, xu2026dahnet}, so most strategies still operate at the local feature level without explicit modeling of global dependencies or adaptive local aggregation for object-specific variations across aerial scenes.

%% file: sec/3_method.tex
\section{Method}
\label{sec:method}

\begin{figure*}
    \centering
    \includegraphics[width=\linewidth]{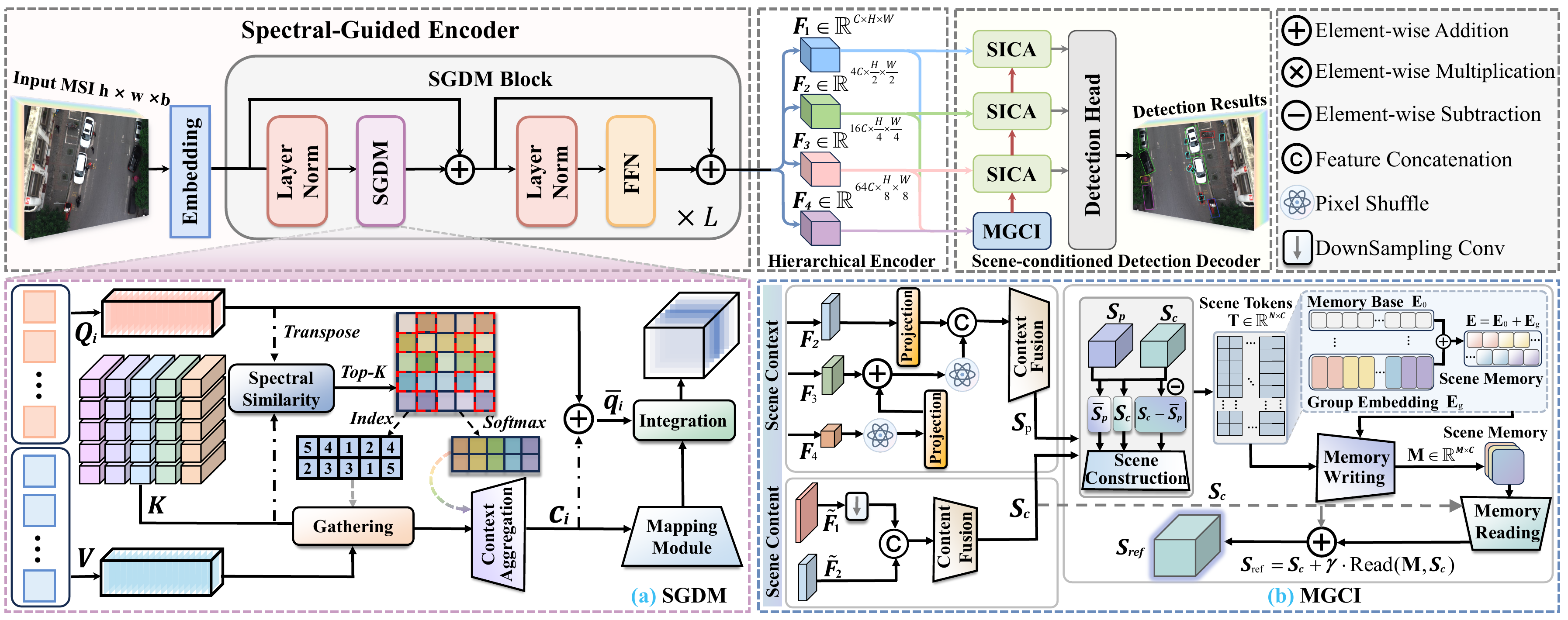}
    \caption{Overall architecture of SGLDet. The spectral-guided encoder first applies SGDM to establish long-range spectral dependencies through sparse retrieval of spectrally similar responses. The scene-conditioned detection decoder then couples MGCI and SICA for hierarchical refinement. (a) SGDM performs sparse spectral retrieval followed by spectral consensus formation and contextual integration. (b) MGCI consolidates distributed cross-scale scene evidence into a compact scene memory and reads out adaptive scene context, while SICA performs scene-dependent adaptive local filtering before detection.}
    \label{fig:framework}
\end{figure*}

We present SGLDet, a spectral-aware global-local dynamic modeling framework for multispectral aerial object detection, as illustrated in~\cref{fig:framework}. SGLDet follows a global-to-local design to jointly address semantic ambiguity and localization challenges in complex aerial scenes. Specifically, the Spectral-driven Global Dynamic Modeling (SGDM) module establishes scene-wide spectral dependencies through similarity-guided sparse retrieval, encouraging semantically consistent responses for spectrally similar objects. The Multi-scale Global Context Integration (MGCI) module then consolidates cross-scale evidence into a compact scene memory and performs scene-guided contextual retrieval for coherent global reasoning. Finally, the Spectral-aware Input-dependent Calibration Aggregation (SICA) module adaptively refines local features through input-dependent expert aggregation and scene-guided spectral-spatial calibration, producing more object-aligned responses for accurate localization.
\subsection{Spectral driven Global Dynamic Modeling}
\label{sec: SGDM}

In multispectral aerial imagery, objects with similar spectral signatures may appear at distant spatial locations, while visually similar regions can still exhibit distinct spectral responses. This property makes purely local interactions inadequate for capturing long-range spectral dependencies across the scene. To this end, we introduce a Spectral-driven Global Dynamic Modeling (SGDM) module, which performs sparse retrieval in the spectral domain using a negative squared Euclidean affinity score, followed by token refinement that promotes spectral consistency and incorporates contextual information.

\noindent\textbf{Token projection and spectral affinity.}
Given an input feature $\boldsymbol{F}\in\mathbb{R}^{C\times H\times W}$, SGDM first transforms it into a token sequence and linearly projects it into query, key, and value embeddings:
\[
[\boldsymbol{Q},\boldsymbol{K},\boldsymbol{V}]=[\phi_{q}(\boldsymbol{F}),\phi_{k}(\boldsymbol{F}),\phi_{v}(\boldsymbol{F})].\tag{1}
\]
where $\boldsymbol{Q},\boldsymbol{K},\boldsymbol{V}\in\mathbb{R}^{N\times C}$ and $N$ denotes the number of pooled spectral tokens. For multi-head modeling, we further partition them as $\boldsymbol{Q}=[\boldsymbol{Q}^{(1)},\dots,\boldsymbol{Q}^{(H)}]$, $\boldsymbol{K}=[\boldsymbol{K}^{(1)},\dots,\boldsymbol{K}^{(H)}]$, and $\boldsymbol{V}=[\boldsymbol{V}^{(1)},\dots,\boldsymbol{V}^{(H)}]$, where $H$ is the number of heads and each head lies in $\mathbb{R}^{N\times d}$ with $C=Hd$. To explicitly model global spectral dependencies, SGDM measures token affinity in each head using the negative squared Euclidean score:
\[
\begin{aligned}
\Psi^{(h)}_{ij}
&=\psi\!\left(\boldsymbol{Q}^{(h)}_{i},\boldsymbol{K}^{(h)}_{j}\right)\\
&=2\left(\boldsymbol{Q}^{(h)}_{i}\right)^{\top}\boldsymbol{K}^{(h)}_{j}-\left\lVert \boldsymbol{Q}^{(h)}_{i}\right\rVert_{2}^{2}-\left\lVert \boldsymbol{K}^{(h)}_{j}\right\rVert_{2}^{2},
\end{aligned}\tag{2}
\]
Equation~(2) is a negative squared Euclidean affinity score, i.e., $\Psi^{(h)}_{ij}=-\lVert\boldsymbol{Q}^{(h)}_{i}-\boldsymbol{K}^{(h)}_{j}\rVert_{2}^{2}$, so a larger score denotes a smaller distance and a more similar query--key pair. For brevity, let $\boldsymbol{q}=\boldsymbol{Q}^{(h)}_{i}$ and $\boldsymbol{k}=\boldsymbol{K}^{(h)}_{j}$, and let $\theta_{ij}$ denote the angle between them. Their squared distance can be decomposed as
\[
\begin{aligned}
\lVert\boldsymbol{q}-\boldsymbol{k}\rVert_{2}^{2}
&=\bigl(\lVert\boldsymbol{q}\rVert_{2}-\lVert\boldsymbol{k}\rVert_{2}\bigr)^{2}\\
&\quad+2\lVert\boldsymbol{q}\rVert_{2}\lVert\boldsymbol{k}\rVert_{2}\bigl(1-\cos\theta_{ij}\bigr).
\end{aligned}
\]
Hence, the negative squared Euclidean score accounts for both angular disagreement and norm mismatch, whereas cosine similarity and Spectral Angle Mapper (SAM) retain only angular information~\cite{kruse1993sips,vandermeer2006similarity}. This distinction is relevant because SGDM applies learned query and key projections without explicit $L_{2}$ normalization after projection. Conversely, if both vectors were $L_{2}$-normalized, Eq.~(2) would reduce to $2\cos\theta_{ij}-2$; because the negative SAM score $-\theta_{ij}=-\arccos(\cos\theta_{ij})$ is strictly increasing in cosine similarity, all three scores would then induce the same Top-$K$ neighbor ordering. We therefore select the negative squared Euclidean score for the unnormalized SGDM feature space and validate this choice empirically in Section~\ref{sec:experiments}.

\noindent\textbf{Top-$K$ sparse spectral retrieval.}
To avoid redundant global interaction, we retain only the Top-$K$ most relevant neighbors for each query token, with $K=\lfloor rN\rfloor$. After masking out all but the Top-$K$ entries in $\boldsymbol{\Psi}^{(h)}_{i,:}$, we set $\widetilde{\Psi}^{(h)}_{i,j}=\Psi^{(h)}_{i,j}$ for retained entries and $-\infty$ otherwise. The resulting sparse retrieval distribution is then given by
\[
\Pi^{(h)}_{i,j}
=
\frac{\exp\!\big(\widetilde{\Psi}^{(h)}_{i,j}\big)}
{\sum_{t=1}^{N}\exp\!\big(\widetilde{\Psi}^{(h)}_{i,t}\big)},
\quad j=1,\dots,N. \tag{3}
\]
The resulting vector $\boldsymbol{\Pi}^{(h)}_{i,:}$ lies on the $(N-1)$-dimensional probability simplex $\Delta^{N-1}=\{\boldsymbol{p}\in\mathbb{R}^{N}\mid p_j\geq 0,\ \sum_{j=1}^{N}p_j=1\}$. Because the Top-$K$ mask assigns zero probability to all unselected neighbors, $\boldsymbol{\Pi}^{(h)}_{i,:}$ is a $K$-sparse point on this simplex. This distribution corresponds to the Top-$K$ spectral retrieval stage in Fig.~\ref{fig:framework}(a), where only the most spectrally compatible neighbors are preserved for subsequent aggregation. After neighbor selection, sparse value aggregation retains $NK$ query--key pairs and scales as
\[
\sum_{i=1}^{N}K=\mathcal{O}(NK),\qquad K\ll N.
\]

\noindent\textbf{Spectral context aggregation.}
Unlike conventional sparse attention, which uses query--key affinity mainly to weight values, SGDM also aggregates selected keys into a spectral context that refines the query before output mapping. Non-local similarity is also used in hyperspectral reconstruction~\cite{zhang2025reconstructive}; in SGDM, learned affinity selects features for object detection. Specifically, the gathered key and value entries corresponding to the Top-$K$ neighbors are aggregated to form a spectral context vector for each query:
\[
\boldsymbol{c}^{(h)}_{i}=\left(\boldsymbol{K}^{(h)}\right)^{\top}\boldsymbol{\Pi}^{(h)}_{i}
+
\beta\left(\boldsymbol{V}^{(h)}\right)^{\top}\boldsymbol{\Pi}^{(h)}_{i}, \tag{4}
\]
where $\beta$ is a learnable coefficient that balances spectral consensus from keys and contextual information from values. The spectral context $\boldsymbol{c}^{(h)}_{i}$ is then used in two complementary paths. First, it is added to the original query to obtain the spectrally refined query $\bar{\boldsymbol{q}}^{(h)}_{i}=\boldsymbol{Q}^{(h)}_{i}+\boldsymbol{c}^{(h)}_{i}$, which aligns each query with its spectrally nearest neighborhood, strengthening global spectral consistency while preserving token identity. Second, the spectral context is passed through a mapping module $\phi_{\mathrm{m}}(\cdot)$ and then integrated with the refined query via residual-scaled injection followed by nonlinear refinement:
\[
\boldsymbol{o}^{(h)}_{i}
=
\mathcal{F}\!\left(\bar{\boldsymbol{q}}^{(h)}_{i}+\gamma\,\phi_{\mathrm{m}}(\boldsymbol{c}^{(h)}_{i})\right), \tag{5}
\]
where $\phi_{\mathrm{m}}(\cdot)$ is a linear projection that transforms the spectral context, $\gamma$ is a learnable scaling coefficient initialized near zero for training stability, and $\mathcal{F}(\cdot)$ denotes a nonlinear feature refinement module that enhances the fused representation. Finally, the responses across all heads are concatenated and projected back to the original embedding space:
\[
\hat{\boldsymbol{F}}_{i}=\phi_{o}\!\left(\operatorname{Concat}\big[\boldsymbol{o}^{(1)}_{i},\dots,\boldsymbol{o}^{(H)}_{i}\big]\right),\tag{6}
\]
where $\phi_o(\cdot)$ denotes the output projection. By routing the spectral context through both the query refinement path and the mapping module before integration, SGDM ensures that the final representation jointly captures spectral coherence and contextual evidence.

Overall, SGDM can be interpreted as a global spectral dependency modeling module for semantic unification rather than a standard dense attention operator. By jointly aggregating key and value information into a unified spectral context and routing it through both query refinement and the final mapping module, SGDM strengthens long-range spectrally guided interaction while preserving token identity for subsequent contextual reasoning. In practice, for the ResNet backbone used in our detector, we instantiate SGDM on the last three backbone stages rather than on the earliest low-level stage, where spectral semantics are less stable and global spectral retrieval is less beneficial.

\subsection{Multi-scale Global Context Integration}
\label{sec: MGCI}
While SGDM establishes spectrally guided long-range dependencies within each pyramid level, reliable recognition in multispectral aerial imagery further requires stable scene-level contextual reasoning, especially when tiny foreground objects are immersed in cluttered backgrounds under severe foreground-background imbalance.
The responses refined by SGDM retain their pyramid levels and spatial identities and therefore remain distributed across scales rather than forming a coherent scene representation.
We therefore introduce a Multi-scale Global Context Integration (MGCI) module, which aligns and consolidates this evidence into a compact scene memory and lets each spatial position retrieve an adaptive context for subsequent local refinement.

\noindent\textbf{Cross-scale scene prior construction.}
Let $\{\boldsymbol{F}_{l}\}_{l=1}^{4}$ denote the hierarchy refined by SGDM. 
MGCI is built upon an intermediate feature scale that balances semantic abstraction and spatial detail. 
At this scale, a scene context $\boldsymbol{S}_{\mathrm{p}}\in\mathbb{R}^{C\times H_{2}\times W_{2}}$ collects higher-level SGDM evidence, where $H_2$ and $W_2$ denote the spatial size aligned with $\boldsymbol{F}_2$:
\[
\boldsymbol{S}_{\mathrm{p}}
= f_{\mathrm{p}}\!\left(
\left[
\phi_{2}(\boldsymbol{F}_{2}),
\mathcal{P}_{s}\!\big(
\boldsymbol{F}_{3}
\,+\,\phi_{4}(\mathcal{P}_{s}(\boldsymbol{F}_{4}))
\big)
\right]
\right).\tag{7}
\label{eq:mgci-prior-content}
\]
Meanwhile, the scene content $\boldsymbol{S}_{\mathrm{c}}\in\mathbb{R}^{C\times H_{2}\times W_{2}}$ preserves spatial detail through $\boldsymbol{S}_{\mathrm{c}}=f_{\mathrm{c}}([\tilde{\boldsymbol{F}}_{2},\mathcal{D}(\tilde{\boldsymbol{F}}_{1})])$, where $\tilde{\boldsymbol{F}}_{1}$ and $\tilde{\boldsymbol{F}}_{2}$ denote the corresponding features after channel alignment. Here, $\mathcal{P}_{s}(\cdot)$ denotes pixel shuffle, $\mathcal{D}(\cdot)$ denotes stride-2 downsampling, and $f_{\mathrm{p}}(\cdot)$ and $f_{\mathrm{c}}(\cdot)$ are fusion mappings. 

\noindent\textbf{Role separation between context and content.}
The context branch summarizes cross-scale scene evidence into a scene context, whereas the content branch preserves the finer spatial detail used to query the compact scene memory. This complementary role separation ensures that MGCI does not simply repeat the spectral retrieval behavior of SGDM, but instead reorganizes distributed scene evidence into a global scene representation for subsequent decoding.

\begin{figure}[t]
    \centering
    \includegraphics[width=\linewidth]{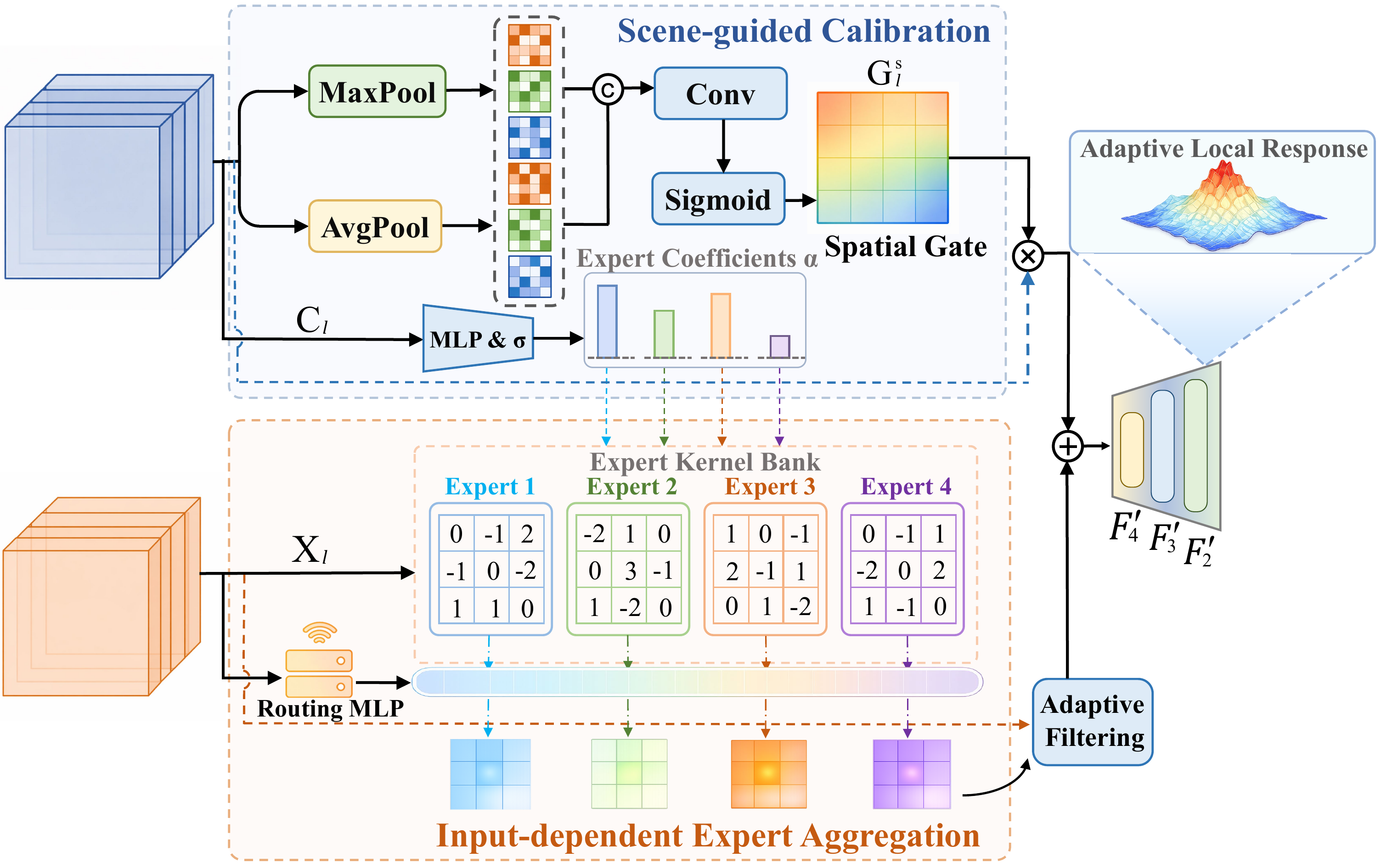}
    \caption{Illustration of the proposed SICA module. SICA performs input-dependent local expert aggregation on $\boldsymbol{X}_{l}$ and scene-guided calibration from $\boldsymbol{C}_{l}$ to adaptively refine multilevel local features.}
    \label{fig:sica}
\end{figure}

\noindent\textbf{Scene token formation and memory initialization.}
Subsequently, MGCI writes distributed scene evidence into compact scene memory and then reads an adaptive scene context from this memory, so scene-level reasoning operates on compact scene memory rather than dense token interactions. We first project the scene context as $\bar{\boldsymbol{S}}_{\mathrm{p}}=\phi_{\mathrm{p}}(\boldsymbol{S}_{\mathrm{p}})$ and then construct scene tokens by coupling the scene content with this projected scene context and their discrepancy at the same decoding scale:

\setcounter{equation}{7}
\begin{equation}
\boldsymbol{T}
= \mathcal{R}\!\left(
f_{\mathrm{s}}\!\big([
\boldsymbol{S}_{\mathrm{c}},
\bar{\boldsymbol{S}}_{\mathrm{p}},
\lvert\boldsymbol{S}_{\mathrm{c}}-\bar{\boldsymbol{S}}_{\mathrm{p}}\rvert
]\big)
\right)
\in \mathbb{R}^{N\times C}.
\tag{8}\label{eq:mgci-scene-tokens}
\end{equation}
where $\mathcal{R}(\cdot)$ denotes adaptive pooling followed by tokenization and $N=H_{s}W_{s}$. The three-way concatenation couples local spatial detail ($\boldsymbol{S}_{\mathrm{c}}$) with the projected scene context ($\bar{\boldsymbol{S}}_{\mathrm{p}}$) and their absolute deviation, so that scene tokens encode both global context and local deviations from it. The $M$ per-image scene slots are initialized by $\boldsymbol{E}=\boldsymbol{E}_{0}+\boldsymbol{E}_{\mathrm{g}}\in\mathbb{R}^{M\times C}$, where $\boldsymbol{E}_{0}$ denotes learnable memory bases and $\boldsymbol{E}_{\mathrm{g}}$ denotes group embeddings. These learned bases initialize a compact scene memory, which is conditioned on the current input through the write operation and used for subsequent context retrieval.

\noindent\textbf{Memory writing.}
We use memory queries together with scene token keys and values, defined as $\boldsymbol{Q}_{\mathrm{w}}=\phi_{\mathrm{w}}^{q}(\boldsymbol{E})$, $\boldsymbol{K}_{\mathrm{w}}=\phi_{\mathrm{w}}^{k}(\boldsymbol{T})$, and $\boldsymbol{V}_{\mathrm{w}}=\phi_{\mathrm{w}}^{v}(\boldsymbol{T})$. The write stage becomes
\begingroup
\setlength{\abovedisplayskip}{4pt plus 1pt minus 1pt}
\setlength{\belowdisplayskip}{4pt plus 1pt minus 1pt}
\setlength{\abovedisplayshortskip}{3pt plus 1pt minus 1pt}
\setlength{\belowdisplayshortskip}{4pt plus 1pt minus 1pt}
\[
\boldsymbol{M}
=
\operatorname{LN}\!\left(
\boldsymbol{E}
\,+\,
\operatorname{MHCA}_{\mathrm{w}}(\boldsymbol{Q}_{\mathrm{w}},\boldsymbol{K}_{\mathrm{w}},\boldsymbol{V}_{\mathrm{w}})
\right)
\in\mathbb{R}^{M\times C},\tag{9}
\label{eq:mgci-write}
\]
\endgroup
where $\operatorname{MHCA}_{\star}(\boldsymbol{Q},\boldsymbol{K},\boldsymbol{V})=\operatorname{Softmax}(\boldsymbol{Q}\boldsymbol{K}^{\top}/\sqrt{d})\boldsymbol{V}$ denotes multi-head cross-attention for slot writing and reading, with $\star\in\{\mathrm{w},\mathrm{r}\}$. The write stage forms $M$ input-conditioned scene slots from $N$ scene tokens and contains $\mathcal{O}(MN)$ query--key score pairs.

\noindent\textbf{Memory reading and contextual refinement.}
For memory reading, scene content queries the compact scene memory to obtain a position adaptive scene context. We instantiate this read step with $\boldsymbol{Q}_{\mathrm{r}}=\phi_{\mathrm{r}}^{q}(\boldsymbol{S}_{\mathrm{c}})$, $\boldsymbol{K}_{\mathrm{r}}=\phi_{\mathrm{r}}^{k}(\boldsymbol{M})$, and $\boldsymbol{V}_{\mathrm{r}}=\phi_{\mathrm{r}}^{v}(\boldsymbol{M})$. The read stage can be formulated as
\begingroup
\setlength{\abovedisplayskip}{4pt plus 1pt minus 1pt}
\setlength{\belowdisplayskip}{4pt plus 1pt minus 1pt}
\setlength{\abovedisplayshortskip}{3pt plus 1pt minus 1pt}
\setlength{\belowdisplayshortskip}{4pt plus 1pt minus 1pt}
\setcounter{equation}{9}
\begin{equation}
\boldsymbol{S}_{\mathrm{ref}}
=
\boldsymbol{S}_{\mathrm{c}}
+
\gamma\,\phi_{\mathrm{o}}\!\left(
\operatorname{MHCA}_{\mathrm{r}}(\boldsymbol{Q}_{\mathrm{r}},\boldsymbol{K}_{\mathrm{r}},\boldsymbol{V}_{\mathrm{r}})
\right).
\tag{10}\label{eq:mgci-read}
\end{equation}
\endgroup
where $\hat{\boldsymbol{S}}=\phi_{\mathrm{o}}(\operatorname{MHCA}_{\mathrm{r}}(\boldsymbol{Q}_{\mathrm{r}},\boldsymbol{K}_{\mathrm{r}},\boldsymbol{V}_{\mathrm{r}}))\in\mathbb{R}^{C\times H_{2}\times W_{2}}$ denotes the retrieved scene context and $\gamma$ is a learnable residual coefficient. Rather than uniformly broadcasting a single global context, this scene-conditioned readout lets each position query the per-image scene slots. Let $P=H_{2}W_{2}$ denote the number of scene-content queries. The read stage contains $\mathcal{O}(PM)$ query--key score pairs, so slot writing and reading together require $\mathcal{O}(MN+PM)=\mathcal{O}(M(N+P))$, compared with $\mathcal{O}(PN)$ for direct MHCA from $P$ queries to $N$ scene tokens. With fixed $M$, this compact representation reduces cross-scale interaction cost while preserving position-specific retrieval from a shared image-conditioned scene state.

\input{sec/table_moda_results}

\subsection{Spectral-aware Input-dependent Calibration Aggregation}
\label{sec: SICA}
While SGDM and MGCI establish global spectral consistency and scene-level contextual coherence, accurate aerial object detection still depends on how local neighborhoods are aggregated. Aerial objects exhibit substantial variation in scale, orientation, aspect ratio, and surrounding clutter, whereas standard convolution applies an identical aggregation pattern to all spatial locations. Such static local modeling is often insufficient for multispectral aerial imagery, where the informative spectral--spatial structure can change significantly across scenes and object instances. To address this issue, we introduce a Spectral-aware Input-dependent Calibration Aggregation (SICA) module, which couples input-dependent local expert aggregation from $\boldsymbol{X}_{l}$ with scene-guided calibration from $\boldsymbol{C}_{l}$ for adaptive local refinement.

\begin{algorithm}[t]
\caption{Pseudo-code of hierarchical SICA refinement}
\label{alg:sica}
\centering
\renewcommand{\arraystretch}{0.92}
\begin{lstlisting}[language={Python}]
# X_pyr: local pyramid features at levels {2, 3, 4}
# S_ref: 256-channel MGCI context at level 2
# W: depth-wise expert bank with M experts
# Gamma[3:4]: separate 3x3 stride-2 convolutions
# (256->256 channels; padding 1; BN + ReLU)
def hierarchical_sica(X_pyr, S_ref, W, Gamma):
    refined = {}
    C = S_ref  # Gamma[2] = Identity
    for level in [2, 3, 4]:
        x_desc = GAP(X_pyr[level])
        alpha = Softmax(route_mlp(x_desc))
        W_mix = sum(alpha[m] * W[m] for m in range(M))
        Z = AdaptiveFiltering(X_pyr[level], W_mix)

        gate_c = Sigmoid(channel_mlp(GAP(C)))
        gate_s = Sigmoid(spatial_conv(cat(AvgPool(C), MaxPool(C))))
        F = out_proj(Z * gate_c + C * gate_s)
        refined[level] = F
        if level < 4:
            C = Gamma[level + 1](F)
    return refined
\end{lstlisting}
\end{algorithm}

As illustrated in Fig.~\ref{fig:sica}, SICA decomposes local refinement into an input-dependent expert aggregation branch and a scene-guided calibration branch. The former produces adaptive local responses from $\boldsymbol{X}_{l}$, while the latter injects scene priors from $\boldsymbol{C}_{l}$ to modulate the routed response.

\noindent\textbf{Input-dependent expert routing.}
Given a local feature map $\boldsymbol{X}_{l}\in\mathbb{R}^{C\times H\times W}$ and its scene guidance $\boldsymbol{C}_{l}\in\mathbb{R}^{C\times H\times W}$, SICA summarizes local content via global average pooling,
\[
\bar{\boldsymbol{x}}_{l}=\operatorname{GAP}(\boldsymbol{X}_{l}). \tag{11}
\label{eq:sica-gap}
\]
Based on this descriptor, SICA predicts routing coefficients over an expert bank,
\[
\alpha_{l,m}=\frac{\exp\!\left(\rho_{m}(\eta_{\mathrm{r}}(\bar{\boldsymbol{x}}_{l}))\right)}{\sum_{t=1}^{M}\exp\!\left(\rho_{t}(\eta_{\mathrm{r}}(\bar{\boldsymbol{x}}_{l}))\right)},\qquad m=1,\dots,M, \tag{12}
\label{eq:sica-alpha}
\]
where $M$ denotes the number of learnable depth-wise experts. Let $\{\boldsymbol{W}_{m}\}_{m=1}^{M}$ be the expert kernel bank with $\boldsymbol{W}_{m}\in\mathbb{R}^{C\times1\times k\times k}$. The routed kernel defines an adaptive filtering operator on $\boldsymbol{X}_{l}$, and the corresponding adaptive local response is written as
\[
\widetilde{\boldsymbol{W}}_{l}=\sum_{m=1}^{M}\alpha_{l,m}\boldsymbol{W}_{m},\qquad
\widetilde{\boldsymbol{Z}}_{l}=\operatorname{DwConv}(\boldsymbol{X}_{l};\widetilde{\boldsymbol{W}}_{l}). \tag{13}
\label{eq:sica-expert}
\]
This formulation preserves the structural prior of convolution while allowing the local aggregation pattern to be adaptively selected from a compact expert bank.

\noindent\textbf{Scene-guided spectral-spatial calibration.}
In parallel, the scene guidance $\boldsymbol{C}_{l}$ is used to calibrate the routed response in complementary channel and spatial manners. The channel gate is estimated from global scene statistics,
\[
\boldsymbol{g}^{\mathrm{c}}_{l}=\sigma\!\left(\psi_{\mathrm{c}}\big(\operatorname{GAP}(\boldsymbol{C}_{l})\big)\right), \tag{14}
\label{eq:sica-channel}
\]
while the spatial gate is predicted from channel-wise pooled scene cues,
\[
\boldsymbol{G}^{\mathrm{s}}_{l}=\sigma\!\left(\psi_{\mathrm{s}}\big([\operatorname{Avg}(\boldsymbol{C}_{l});\operatorname{Max}(\boldsymbol{C}_{l})]\big)\right). \tag{15}
\label{eq:sica-spatial}
\]
where $\operatorname{Avg}(\cdot)$ and $\operatorname{Max}(\cdot)$ denote channel-wise average and max pooling. The final refined response is obtained by projecting the scene-calibrated local response,
\[
\boldsymbol{F}_{l}^{\prime}=\phi_{\mathrm{o}}\!\left(\widetilde{\boldsymbol{Z}}_{l}\odot\boldsymbol{g}^{\mathrm{c}}_{l}+\boldsymbol{G}^{\mathrm{s}}_{l}\odot\boldsymbol{C}_{l}\right), \tag{16}
\label{eq:sica-output}
\]
where $\phi_{\mathrm{o}}(\cdot)$ denotes the output projection. Algorithm~\ref{alg:sica} summarizes the hierarchy. Here, $\Gamma_2=\mathrm{Id}$, whereas $\Gamma_3$ and $\Gamma_4$ are independent $3\times3$ convolutions with stride 2 and padding 1, followed by batch normalization and ReLU. In practice, SICA is applied from shallow to deep levels, passing scene-aware refinement to higher-level features. Consequently, its adaptive local filtering complements the global reasoning of SGDM and MGCI with content-aware refinement.

%% file: sec/table_moda_results.tex
\begin{table*}[!t]
\centering
\small
\caption{Comparison with other methods on MODA. $^\dagger$ indicates methods originally designed for multispectral object detection.}
\setlength{\tabcolsep}{3.0pt}
\begin{tabular}{l|cccccccc|ccc|cc}
\toprule
\textbf{Method} & \textbf{car} & \textbf{bus} & \textbf{van} & \textbf{aw-bike} & \textbf{truck} & \textbf{tricycle} & \textbf{bike} & \textbf{pedestrian} & \textbf{mAP$_{50}\uparrow$} & \textbf{mAP$_{75}\uparrow$} & \textbf{mAP$\uparrow$} & \textbf{FLOPs$\downarrow$} & \textbf{Params$\downarrow$} \\
\midrule
\multicolumn{14}{c}{\textbf{two-stage methods}} \\
\midrule
Roi Transformer \cite{roi_trans}    & 90.5 & 89.3 & 75.2 & 73.3 & 68.7 & 51.8 & 44.5 & 30.0 & 65.4 & 43.4 & 40.7 & 244.7G & 55.3M \\
O-RCNN \cite{xie2021oriented}       & 90.5 & 89.7 & 74.6 & 72.5 & 66.6 & 54.7 & 45.1 & 30.4 & 65.5 & 44.0 & 40.9 & 244.7G & 55.3M \\
LSKNet \cite{Li_2024_IJCV}          & 90.4 & 88.5 & 74.5 & 72.5 & 66.0 & 52.5 & 41.0 & 28.2 & 64.2 & 42.8 & 39.4 & 221.9G & 32.6M \\
StripRCNN \cite{yuan2025strip}      & 90.5 & 89.0 & 76.1 & 73.6 & 67.1 & 56.5 & 45.2 & 30.7 & 66.1 & 44.0 & 41.0 & 231.8G & 45.2M \\
ReDet \cite{redet}                  & 90.6 & 89.6 & 76.8 & 72.9 & 69.6 & 52.4 & 41.5 & 41.0 & 66.8 & 44.1 & 41.3 & 245.2G & 31.6M \\
\midrule
\multicolumn{14}{c}{\textbf{one-stage methods}} \\
\midrule
Rot. Retina \cite{retinanet}        & 90.2 & 81.9 & 72.4 & 59.0 & 45.5 & 30.8 & 21.7 & 25.4 & 53.4 & 37.3 & 34.1 & 233.5G & 36.5M \\
R3Det \cite{yang2021r3det}          & 90.4 & 88.7 & 74.0 & 63.8 & 57.8 & 45.1 & 29.4 & 32.2 & 60.2 & 36.7 & 34.8 & 362.2G & 42.0M \\
S$^2$ANet \cite{han2021align}       & 90.4 & 88.7 & 74.4 & 69.1 & 62.7 & 48.9 & 30.1 & 40.7 & 63.1 & 38.7 & 36.7 & 216.5G & 38.8M \\
S2ADet$^\dagger$ \cite{he2023object}       & 90.3 & 86.6 & 72.1 & 71.3 & 57.2 & 54.1 & 35.0 & 40.8 & 63.5 & 41.1 & 38.9 & 406.0G & 65.2M \\
Rot. FCOS \cite{tian2019fcos}               & 90.3 & 86.3 & 75.2 & 71.2 & 59.0 & 53.8 & 34.7 & 37.4 & 63.5 & 41.8 & 39.1 & 226.9G & 32.2M \\
GWD \cite{yang2021rethinking}       & 90.4 & 79.9 & 70.7 & 59.7 & 49.3 & 30.8 & 26.7 & 29.7 & 54.7 & 37.8 & 34.6 & 233.5G & 36.5M \\
RepPts-R \cite{reppoints}           & 89.9 & 70.2 & 67.0 & 66.1 & 50.2 & 41.5 & 33.3 & 40.5 & 57.3 & 34.9 & 33.0 & \textbf{213.6G} & 36.9M \\
R3Det-KLD \cite{yang2021learning}   & 90.4 & 88.7 & 74.8 & 66.6 & 60.9 & 42.6 & 29.8 & 35.5 & 61.2 & 39.4 & 36.8 & 306.7G & 39.6M \\
CFA \cite{cfa}                      & 90.4 & 89.0 & 76.7 & 69.7 & 64.4 & 55.1 & 43.1 & 41.5 & 66.2 & 43.2 & 40.6 & \textbf{213.6G} & 36.9M \\
Ori. RepPoints \cite{oriented-reppoints} & 90.5 & 89.2 & 77.7 & 71.2 & 66.2 & 53.1 & 43.0 & 41.1 & 66.5 & 44.1 & 40.9 & \textbf{213.6G} & 36.9M \\
RTMDet \cite{lyu2022rtmdet}         & 90.0 & 88.4 & 75.3 & 71.1 & 62.8 & 52.5 & 42.9 & 41.0 & 65.5 & 41.5 & 40.1 & 258.5G & 52.3M \\
DCFL \cite{Xu_2023_CVPR}            & 90.6 & 89.7 & 77.5 & 71.3 & 67.2 & 55.5 & 44.2 & 44.0 & 67.5 & 44.8 & 41.5 & 241.0G & 37.0M \\
MessDet \cite{wu2025measuring}      & 90.0 & 89.1 & 75.4 & 71.4 & 70.2 & 54.9 & 48.4 & 45.5 & 68.1 & 42.4 & 40.1 & 616.0G & \textbf{15.1M} \\
OSSDet$^\dagger$ \cite{han2025moda} & 90.5 & 89.9 & 79.2 & 72.7 & 69.7 & 58.8 & 45.3 & 45.7 & 69.0 & 45.9 & 42.7 & 263.1G & 36.5M \\
\midrule
\multicolumn{14}{c}{\textbf{DETR-based methods}} \\
\midrule
AO$^2$-DETR \cite{dai2022ao2}       & 89.4 & 88.1 & 76.0 & 69.5 & 65.7 & 54.8 & 41.9 & 41.0 & 65.8 & 43.0 & 40.0 & 240.0G & 42.0M \\
ARS-DETR \cite{zeng2024ars}          & 90.5 & 89.2 & 77.2 & 70.6 & 64.8 & 53.6 & 43.2 & 39.7 & 66.1 & 43.4 & 40.5 & 254.0G & 44.2M \\
RHINO \cite{Lee_2025_WACV}           & 90.7 & 89.1 & 77.3 & 70.8 & 65.6 & 54.0 & 42.9 & 41.0 & 66.4 & 44.0 & 40.8 & 319.5G & 50.8M \\
Ori-DETR \cite{zhao2024pointaxis}    & \textbf{90.8} & 89.3 & 78.8 & 72.8 & 68.0 & 56.0 & 46.6 & 40.1 & 67.8 & 45.1 & 41.9 & 492.0G & 57.6M \\
\midrule
\rowcolor{lightpurple} \textbf{SGLDet (Ours)} & 90.5 & \textbf{90.2} & \textbf{79.3} & \textbf{75.8} & \textbf{70.8} & \textbf{65.6} & \textbf{51.5} & \textbf{50.9} & \textbf{71.8} & \textbf{47.0} & \textbf{44.3} & 241.6G & 37.1M \\
\bottomrule
\end{tabular}
\label{tab2}
\vspace{-2mm}
\end{table*}

%% file: sec/4_experiments.tex
\section{Experiments}
\label{sec:experiments}

\looseness=1
We evaluate SGLDet on three benchmarks with different input configurations and box annotations, including multiband images that retain wavelength-resolved information~\cite{huang2022spectral,williams2025hyperpixels}. MODA~\cite{han2025moda} uses eight-band images with oriented boxes; HOD3K~\cite{he2023object} uses 16-band images with horizontal boxes; and DroneVehicle~\cite{dronevehicle} provides paired RGB--infrared images with oriented boxes. The MODA and HOD3K comparisons therefore follow oriented- and horizontal-box evaluation protocols, respectively. Within these settings, MODA covers two-stage, one-stage, and DETR-based oriented detectors alongside multispectral methods; HOD3K covers horizontal-box detectors; and DroneVehicle includes single-modality and RGB--infrared fusion methods. We report mAP$_{50}$ and mAP on all three benchmarks. For MODA and HOD3K, we additionally report mAP$_{75}$. We also report FLOPs, model parameters, and hardware-controlled wall-clock measurements. Comparison results are taken from prior studies or obtained in this work. Component ablations and detector compatibility studies are conducted on MODA; the input-modality and fusion comparison is conducted on DroneVehicle.

Unless otherwise specified, all SGLDet training experiments and ablation variants were implemented in MMRotate with PyTorch 1.11.0 and conducted on two NVIDIA RTX 3090 GPUs (48 GB total). SGLDet builds on Oriented RepPoints~\cite{oriented-reppoints} while preserving the original detection head. ResNet-50 pretrained on ImageNet was adopted as the backbone, and AdamW was used with a weight decay of 0.03 and gradient clipping with a maximum norm of 25. For each benchmark, a separate model was trained using its native multichannel input. Only the first ResNet-50 convolution was randomly initialized; the remaining backbone layers inherited ImageNet-pretrained weights. The ResNet-50 input stem is adapted to each dataset's channel count while preserving its output width. The three-module configuration (SGDM+MGCI+SICA) follows the same feature-processing flow across the three input settings.

\begin{figure*}
    \centering
    \includegraphics[width=\linewidth]{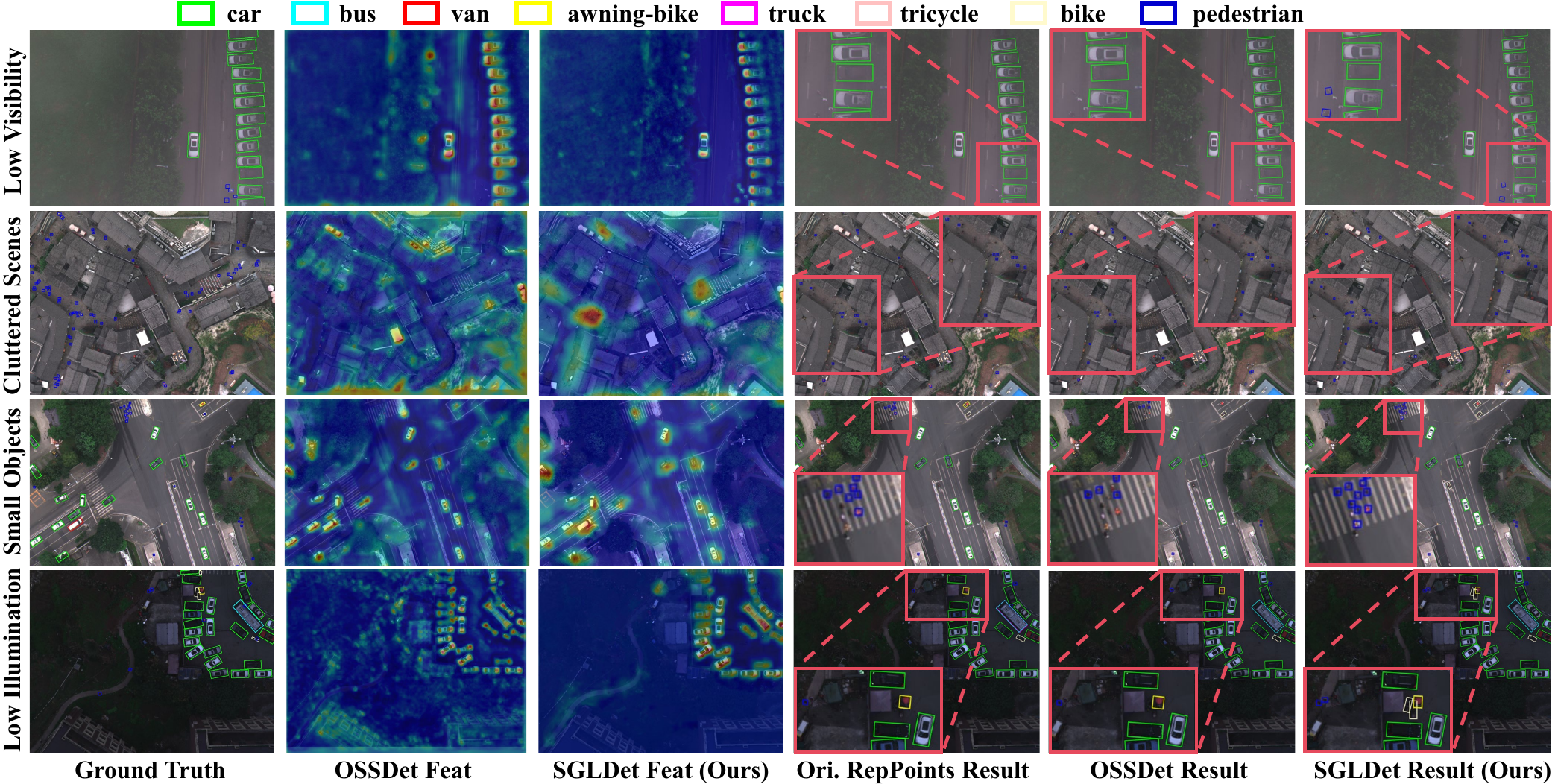}
    \caption{Visualization comparison of detection results and feature maps obtained by different methods on MODA.}
    \label{fig:MODA_results_vis}
\end{figure*}
\begin{figure}[t]
    \centering
    \includegraphics[width=\linewidth]{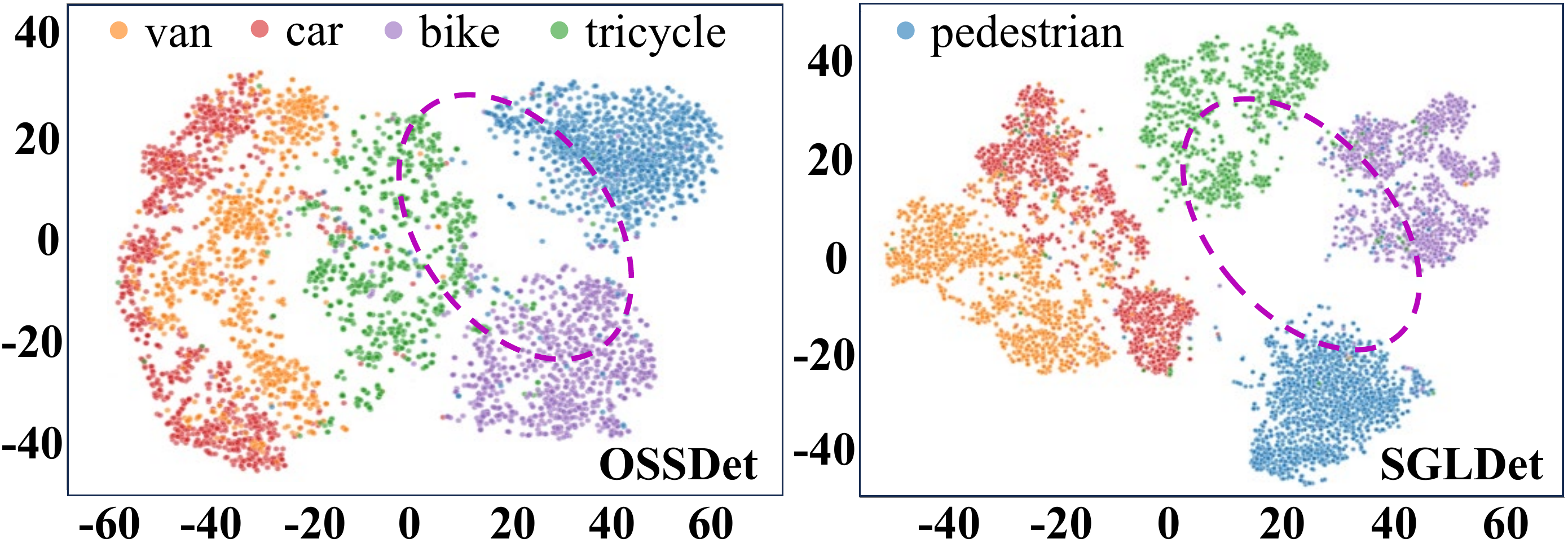}
    \caption{Two-dimensional t-SNE visualization of feature distributions learned by OSSDet and SGLDet on MODA.}
    \label{fig:tsne}
\end{figure}
\begin{figure}[t]
    \centering
    \includegraphics[width=\linewidth]{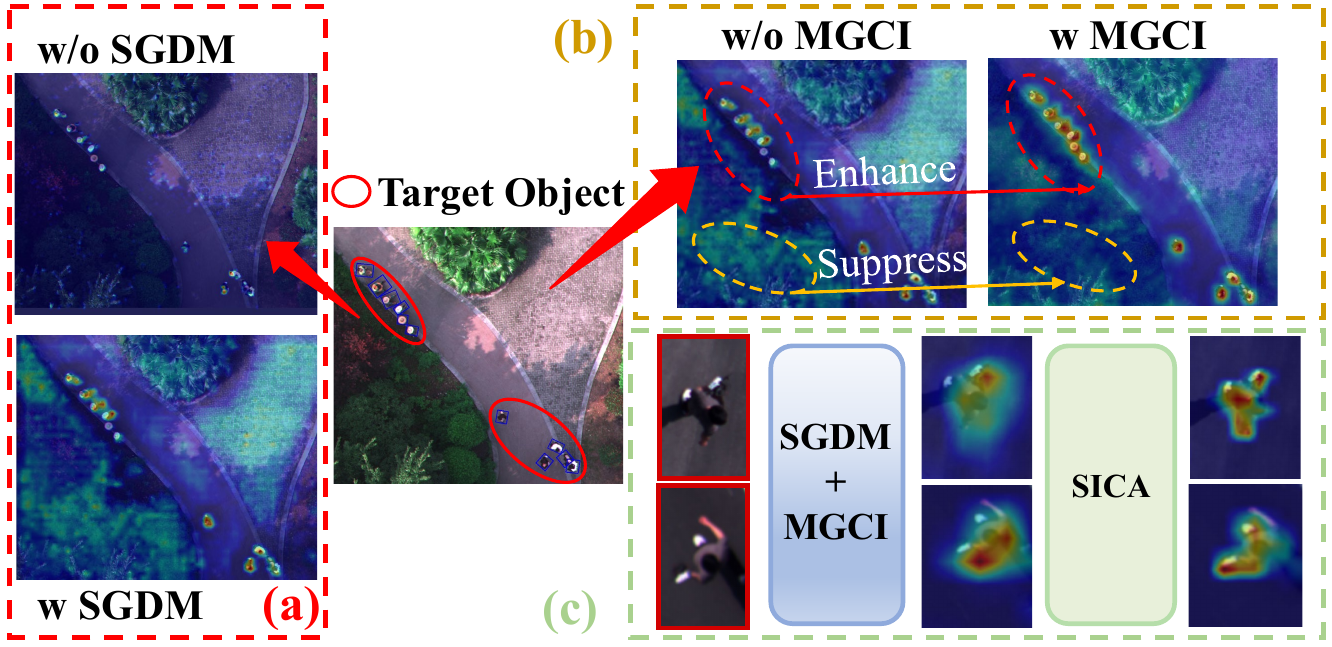}
    \caption{Effectiveness visualization of key components.}
    \label{fig:sacf_feature_map}
\end{figure}
\subsection{Results on the MODA Dataset}
\noindent\textbf{Data and Setups.}
The MODA dataset~\cite{han2025moda} is a large-scale multispectral aerial benchmark with 14,041 images and 330,191 oriented annotations over 8 categories. Each image has a resolution of $1200 \times 900$ and covers 8 spectral bands from approximately 395 to 950 nm. MODA contains diverse urban scenes with small objects, low illumination, occlusion, truncation, and cluttered backgrounds. Each image was normalized and zero-padded from $1200\times900$ to $1216\times928$. The model was trained for 12 epochs with a total batch size of 4. The learning rate was initialized to $7\times10^{-5}$ and scheduled by cosine annealing with 700 linear warm-up iterations, a warm-up ratio of $1/3$, and a minimum learning rate of $7\times10^{-7}$.

\noindent\textbf{Quantitative Results.} \cref{tab2} lists the quantitative results of state-of-the-art detectors and our methods on the MODA dataset. SGLDet achieves the best overall detection performance and outperforms the strongest MSI-oriented baseline, OSSDet, by 2.8\%, 1.1\%, and 1.6\% in mAP$_{50}$, mAP$_{75}$, and mAP, respectively. In class-wise evaluations, SGLDet achieves the best results in seven out of eight categories, with the largest gain of 6.8\% on ``tricycle'', showing the effectiveness of spectral-cue-guided global and local dynamic modeling for multispectral aerial object detection.

\noindent\textbf{Performance on Challenging Subsets.} To construct three fixed challenging subsets, we compute three model-independent scores for each test image $i$:
\setcounter{equation}{16}
\begin{equation}
\begin{aligned}
L_i &= \frac{1}{8}\sum_{c=1}^{8}\left\langle \boldsymbol{X}_{i,c}\right\rangle_p,\\[-2pt]
S_i &= \underset{1\leq j\leq n_i}{\operatorname{median}}\sqrt{A_{i,j}}, \qquad D_i=n_i,
\end{aligned}
\end{equation}
where $\boldsymbol{X}_{i,c}$ is the $c$th radiometrically corrected native spectral band before model normalization and padding, $\langle\cdot\rangle_p$ denotes its spatial mean, and $A_{i,j}$ is the annotated oriented-box area. Thus, $L_i$ averages the spatial means of all eight bands, $S_i$ is the per-image median ground-truth object scale, and $D_i$ is the instance count. Every image in the 4,885-image MODA test set contains at least one valid annotated instance, so $S_i$ is defined for every image. We rank all 4,885 images independently by $L_i$, $S_i$, and $D_i$ with filename tie-breaking. The lowest 20\%, lowest 20\%, and highest 20\% form the low-illumination, small-object, and high-density subsets, respectively (977 images each). Across these subsets, SGLDet outperforms OSSDet, with its largest gains under low illumination (3.7\% mAP$_{50}$ and 2.5\% mAP).

\input{sec/table_moda_challenging_subsets_main}

\noindent\textbf{Efficiency Analysis.} As~\cref{tab2} shows, two-stage methods offer high accuracy but complex architectures, while conventional one-stage detectors are efficient yet less accurate. The MSI-oriented S2ADet adopts a two-stream design to decouple spatial and spectral cues, which incurs substantial preprocessing and computational cost. In contrast, SGLDet achieves the highest mAP$_{50}$ with competitive computational cost, striking a favorable accuracy-efficiency balance.

\noindent\textbf{Qualitative Results.}
As depicted in \cref{fig:MODA_results_vis}, our method enhances the network’s focus on target regions while suppressing irrelevant background noise, even when objects are small or captured under low-visibility, low-illumination, and cluttered conditions. Fewer false positives and missed detections are observed. These improvements are consistent with the integration of spectrally guided global retrieval and input-adaptive local modulation, which jointly enhance the classification and localization of small objects with similar spectral signatures in complex backgrounds.

\noindent\textbf{Visualization of Feature Distribution.}
T-SNE~\cite{van2008visualizing} visualization results are shown in~\cref{fig:tsne}. Compared with OSSDet, the features extracted by SGLDet form more compact clusters within the same category, indicating enhanced intra-class consistency. Meanwhile, SGLDet shows wider inter-class separation for easily confused categories (e.g., car and van), small objects (e.g., pedestrian and bike), and few-sample categories (e.g., tricycle). This visualization supports the role of spectral-driven global dynamic modeling in making spectrally similar objects more semantically coherent while improving inter-class discriminability through adaptive global--local modulation.

\noindent\textbf{Effect of Components.} \cref{fig:sacf_feature_map} visualizes features involved in key components of SGLDet. (a) SGDM reinforces semantic coherence among spectrally similar objects through global retrieval of the most relevant spectral responses, thereby enhancing the responses of such objects across the scene. (b) MGCI integrates multi-scale features and establishes holistic scene perception through global contextual reasoning, strengthening the contrast between foreground objects and background regions. (c) SICA performs scene-guided local expert aggregation to adaptively refine local responses, enabling more discriminative local aggregation for fine-grained detection.

\subsection{Ablation Studies and Analysis}
To evaluate the practical efficiency, key components, and compatibility of SGLDet with existing detectors, we conduct a series of analysis and ablation experiments on the MODA dataset. Unless otherwise specified, all studies follow the same MODA setting used in the main comparison.

\noindent\textbf{Measured Inference Efficiency.}
\cref{tab:inference-efficiency} reports detection accuracy, FLOPs, inference latency, and FPS. All methods are benchmarked on a single idle NVIDIA RTX 3090 GPU with batch size 1, FP32 inference, and no test-time augmentation. Each detector uses the test-time post-processing configuration adopted for its MODA accuracy evaluation, including the score threshold and NMS settings where applicable. Native $1200\times900$ inputs are normalized and padded to $1216\times928$. Latency includes network forward propagation, prediction decoding, and NMS where applicable, while excluding I/O and memory-copy operations. After 50 warm-up iterations, all detectors are timed on the same fixed subset of 500 MODA test images in each of three independent runs. CUDA is synchronized at both timing boundaries; latency is reported as the mean $\pm$ standard deviation of the three run means, and FPS is the reciprocal of the overall mean latency. SGLDet runs at 54~ms per image (18.5 FPS). It is 2~ms faster than Ori.~RepPoints (56~ms, 17.9 FPS) while improving mAP$_{50}$ by 5.3 percentage points. Compared with OSSDet, SGLDet incurs a marginal 1-ms latency increase (1.9\%) while improving mAP$_{50}$ and mAP by 2.8 and 1.6 percentage points, respectively. SGDM requires Top-$K$ neighbor selection and feature gathering, whose runtime costs are not directly reflected in FLOPs; fewer FLOPs need not imply lower latency~\cite{pan2022hilo}. This controlled comparison shows that SGLDet achieves a favorable overall balance between detection accuracy and measured inference efficiency.

\input{sec/table_inference_efficiency_compact}

\noindent\textbf{Component Ablation Study.}
As shown in \cref{tab5}, SGDM and MGCI individually improve mAP$_{50}$ from 66.5\% to 68.4\% and 68.0\%, respectively, while their combination reaches 70.2\%. SGDM strengthens within-scale spectral consistency, whereas MGCI organizes the refined responses into cross-scale scene context. Adding SICA further raises mAP$_{50}$ to 71.8\%, yielding a 5.3\% improvement over the baseline and demonstrating the synergy among spectrally guided global retrieval, scene-level contextual reasoning, and input-adaptive local aggregation.

\begin{table}[!t]
    \centering
    \small
    \caption{Ablation studies on the key components of SGLDet.}
    \setlength{\tabcolsep}{5.5pt}
    \begin{tabular}{ccc|ccc}
        \toprule
SGDM & MGCI & SICA & mAP$_{50}\uparrow$ & mAP$_{75}\uparrow$ & mAP$\uparrow$ \\
        \midrule
        \xmark & \xmark & \xmark & 66.5 & 44.1 & 40.9 \\
        \cmark & \xmark  & \xmark & 68.4 & 44.5 & 41.4 \\
        \xmark & \cmark  & \xmark & 68.0 & 44.6 & 41.5 \\
        \xmark & \xmark  & \cmark & 68.3 & 44.8 & 41.7 \\
        \cmark & \cmark  & \xmark & 70.2 & 45.8 & 42.6 \\
        \cmark & \cmark  & \cmark  & \textbf{71.8} & \textbf{47.0} & \textbf{44.3} \\
        \bottomrule
    \end{tabular}
    \label{tab5}
\end{table}

\noindent\textbf{Ablation Study on SGDM.}
A key design in SGDM is the spectral consensus mechanism, which aligns each query with its spectrally nearest neighborhood before contextual aggregation. As shown in \cref{tab:consensus}, removing this consensus step and directly aggregating retrieved values leads to a 0.6\% drop in mAP$_{50}$ and a 1.0\% drop in mAP, confirming that explicit query-side consensus formation strengthens spectral consistency and is beneficial for subsequent content retrieval. We further evaluate the Top-$K$ selection ratio $r$ that controls the sparsity of spectral retrieval. As shown in \cref{tab:topk-ratio}, a small $r$ restricts global interaction and may fail to capture sufficient spectrally consistent cues, while a large $r$ introduces redundant or irrelevant features that dilute the spectral specificity of target tokens. SGLDet achieves the best results when $r=0.6$, striking a balance between retrieval coverage and noise suppression.

We compare the negative squared Euclidean score with cosine similarity $\Psi^{\mathrm{cos}}_{ij}=\boldsymbol{Q}_{i}^{\top}\boldsymbol{K}_{j}/(\lVert\boldsymbol{Q}_{i}\rVert_{2}\lVert\boldsymbol{K}_{j}\rVert_{2})$ and negative SAM $\Psi^{\mathrm{SAM}}_{ij}=-\arccos(\operatorname{clip}(\Psi^{\mathrm{cos}}_{ij},-1+\epsilon,1-\epsilon))$, where $\epsilon$ is a small numerical constant and $\operatorname{clip}(\cdot)$ bounds its input to the displayed interval for numerical stability. All other settings are fixed, and larger scores indicate closer neighbors in every variant. As shown in \cref{tab:sgdm-affinity}, the negative squared Euclidean score achieves the highest mAP of 44.3\%, outperforming cosine similarity and negative SAM by 0.8\% and 1.2\%, respectively. These results support its use as SGDM's affinity function.

\begin{table}[!t]
\centering
\small
\caption{Ablation study of the Top-$K$ ratio in SGDM.}
\setlength{\tabcolsep}{16pt}
    \begin{tabular}{c|ccc}
    \toprule
$r$ & mAP$_{50}\uparrow$ & mAP$_{75}\uparrow$ & mAP$\uparrow$\\
    \midrule
    0.2 & 70.6 & 45.7 & 42.5 \\
    0.4 & 71.4 & 46.4 & 43.2 \\
    \rowcolor{lightpurple}0.6 & \textbf{71.8 }& \textbf{47.0} & \textbf{44.3} \\
    0.8 & 71.0 & 46.1 & 42.9 \\
    \bottomrule
\end{tabular}
\label{tab:topk-ratio}
\end{table}

\input{sec/table_sgdm_affinity}
\begin{table*}[!t]
\begin{minipage}[t]{0.48\textwidth}
    \centering
    \small
    \caption{Ablation study on spectral consensus in SGDM.}
    \setlength{\tabcolsep}{7pt}
    \begin{tabular}{c|ccc}
    \toprule
Variant & mAP$_{50}\uparrow$ & mAP$_{75}\uparrow$ & mAP$\uparrow$\\
    \midrule
    w/o spectral consensus & 71.2 & 46.5 & 43.3 \\
    \rowcolor{lightpurple}w/ spectral consensus & \textbf{71.8} & \textbf{47.0} & \textbf{44.3} \\
    \bottomrule
    \end{tabular}
    \label{tab:consensus}
\end{minipage}
\hfill
\begin{minipage}[t]{0.48\textwidth}
    \centering
    \small
    \caption{Ablation study of the affinity function in SGDM.}
    \setlength{\tabcolsep}{3.5pt}
    \sgdmaffinitytabular
    \label{tab:sgdm-affinity}
\end{minipage}
\end{table*}

\noindent\textbf{Ablation Study on MGCI.}
MGCI uses a slot-writing stage to consolidate cross-scale tokens and a context-readout stage to retrieve position-adaptive scene context. \Cref{tab:mgci-slots} first studies the number of scene slots. Increasing $M$ from 8 to 16 improves all three metrics, whereas using 32 slots causes a slight decline. This suggests that too few slots limit contextual capacity, while too many may introduce redundant scene representations. We therefore use $M=16$ to balance context capacity and memory compactness. \Cref{tab:mgci-design} then compares the complete slot write--read path with a direct-MHCA variant that removes slot writing and lets the $P$ scene-content queries attend directly to all $N$ scene tokens. Compared with direct MHCA, the complete design improves mAP$_{50}$, mAP$_{75}$, and mAP by 0.6\%, 0.6\%, and 1.2\%, respectively, supporting the use of compact scene memory for cross-scale context aggregation.

\begin{table*}[!t]
\begin{minipage}[t]{0.48\textwidth}
    \centering
    \small
    \caption{Ablation study of the scene-memory cardinality in MGCI.}
    \setlength{\tabcolsep}{17pt}
    \begin{tabular}{c|ccc}
    \toprule
    $M$ & mAP$_{50}\uparrow$ & mAP$_{75}\uparrow$ & mAP$\uparrow$ \\
    \midrule
    8 & 71.3 & 46.2 & 43.1 \\
    \rowcolor{lightpurple}16 & \textbf{71.8} & \textbf{47.0} & \textbf{44.3} \\
    32 & 71.5 & 46.6 & 43.4 \\
    \bottomrule
    \end{tabular}
    \label{tab:mgci-slots}
\end{minipage}
\hfill
\begin{minipage}[t]{0.48\textwidth}
    \centering
    \small
    \caption{Ablation study on the key design choices of MGCI.}
    \setlength{\tabcolsep}{2.0pt}
    \begin{tabular}{cc|ccc}
    \toprule
    Compact Memory & Scene Readout & mAP$_{50}\uparrow$ & mAP$_{75}\uparrow$ & mAP$\uparrow$ \\
    \midrule
    \xmark & \xmark & 70.3 & 45.7 & 42.6 \\
    \cmark & \xmark & 71.0 & 46.2 & 43.0 \\
    \xmark & \cmark & 71.2 & 46.4 & 43.1 \\
    \cmark & \cmark & \textbf{71.8} & \textbf{47.0} & \textbf{44.3} \\
    \bottomrule
    \end{tabular}
    \label{tab:mgci-design}
\end{minipage}
\end{table*}

\noindent\textbf{Ablation Study on SICA.}
SICA performs scene-guided local expert aggregation through a learnable expert bank together with lightweight channel-spatial calibration. We conduct two groups of experiments to validate these design choices. First, \cref{tab:sica-design} examines the contribution of expert diversity and scene-guided calibration. Replacing the expert bank with a single expert ($M{=}1$) reduces mAP$_{50}$ by 1.4\% and mAP by 1.6\%, indicating that a shared local filter is insufficient to capture the heterogeneous local patterns of aerial objects. Removing channel calibration or spatial calibration also degrades performance, confirming that global scene cues provide complementary guidance for adaptive local refinement beyond the routed expert response alone.
We further evaluate the number of experts in \cref{tab:sica-prototypes}.
Using too few experts limits the diversity of local aggregation patterns, whereas an excessively large expert bank introduces redundancy and slightly weakens generalization. The best trade-off is obtained with $M{=}8$, which is therefore used in all experiments.

\begin{table*}[!t]
\begin{minipage}[t]{0.48\textwidth}
    \centering
    \small
    \caption{Ablation study on the key design choices of SICA.}
    \setlength{\tabcolsep}{6.5pt}
    \begin{tabular}{c|ccc}
    \toprule
    Variant & mAP$_{50}\uparrow$ & mAP$_{75}\uparrow$ & mAP$\uparrow$\\
    \midrule
    Single Expert ($M{=}1$) & 70.4 & 45.8 & 42.7 \\
    w/o Channel Calibration & 71.3 & 46.5 & 43.4 \\
    w/o Spatial Calibration & 71.1 & 46.3 & 43.2 \\
    \rowcolor{lightpurple}Full SICA & \textbf{71.8} & \textbf{47.0} & \textbf{44.3} \\
    \bottomrule
    \end{tabular}
    \label{tab:sica-design}
\end{minipage}
\hfill
\begin{minipage}[t]{0.48\textwidth}
    \centering
    \small
    \caption{Ablation study of the number of experts in SICA.}
    \setlength{\tabcolsep}{16.5pt}
    \begin{tabular}{c|ccc}
    \toprule
    $M$ & mAP$_{50}\uparrow$ & mAP$_{75}\uparrow$ & mAP$\uparrow$\\
    \midrule
    4 & 70.3 & 45.8 & 42.7 \\
    \rowcolor{lightpurple}8 & \textbf{71.8} & \textbf{47.0} & \textbf{44.3} \\
    12 & 71.4 & 46.7 & 43.5 \\
    16 & 71.2 & 46.5 & 43.3 \\
    \bottomrule
    \end{tabular}
    \label{tab:sica-prototypes}
\end{minipage}
\end{table*}

\noindent\textbf{Compatibility Analysis.}
To assess compatibility with existing detectors, we integrate SGDM into the last three stages of the R50 backbone in five representative models; the baseline rows are taken from \cref{tab2}. As shown in \cref{tab:compatibility-moda}, SGDM improves every baseline, with mAP$_{50}$ gains ranging from 1.1\% to 1.6\%. The gains across different detector families support its use as a broadly compatible backbone enhancement rather than a head-specific modification.

\begin{table}[!t]
\centering
\small
\caption{Compatibility analysis on representative R50-based detectors on MODA. Baseline results are taken from \cref{tab2}.}
\setlength{\tabcolsep}{3pt}
\begin{tabular}{l|c|lll}
\toprule
\textbf{Detector} & \textbf{Backbone} & \textbf{mAP$_{50}\uparrow$} & \textbf{mAP$_{75}\uparrow$} & \textbf{mAP$\uparrow$} \\
\midrule
\multirow{2}{*}{O-RCNN} & R50 & 65.5 & 44.0 & 40.9 \\
 & \cellcolor{lightpurple}\textbf{R50$_{\text{SGDM}}$} & \cellcolor{lightpurple}\textbf{67.1}(+1.6) & \cellcolor{lightpurple}\textbf{45.2}(+1.2) & \cellcolor{lightpurple}\textbf{41.9}(+1.0) \\
\midrule
\multirow{2}{*}{Rot. FCOS} & R50 & 63.5 & 41.8 & 39.1 \\
 & \cellcolor{lightpurple}\textbf{R50$_{\text{SGDM}}$} & \cellcolor{lightpurple}\textbf{64.9}(+1.4) & \cellcolor{lightpurple}\textbf{42.9}(+1.1) & \cellcolor{lightpurple}\textbf{40.0}(+0.9) \\
\midrule
\multirow{2}{*}{CFA} & R50 & 66.2 & 43.2 & 40.6 \\
 & \cellcolor{lightpurple}\textbf{R50$_{\text{SGDM}}$} & \cellcolor{lightpurple}\textbf{67.7}(+1.5) & \cellcolor{lightpurple}\textbf{44.3}(+1.1) & \cellcolor{lightpurple}\textbf{41.6}(+1.0) \\
\midrule
\multirow{2}{*}{Ori. RepPoints} & R50 & 66.5 & 44.1 & 40.9 \\
 & \cellcolor{lightpurple}\textbf{R50$_{\text{SGDM}}$} & \cellcolor{lightpurple}\textbf{67.8}(+1.3) & \cellcolor{lightpurple}\textbf{45.1}(+1.0) & \cellcolor{lightpurple}\textbf{41.8}(+0.9) \\
\midrule
\multirow{2}{*}{Ori-DETR} & R50 & 67.8 & 45.1 & 41.9 \\
 & \cellcolor{lightpurple}\textbf{R50$_{\text{SGDM}}$} & \cellcolor{lightpurple}\textbf{68.9}(+1.1) & \cellcolor{lightpurple}\textbf{45.9}(+0.8) & \cellcolor{lightpurple}\textbf{42.7}(+0.8) \\
\bottomrule
\end{tabular}
\label{tab:compatibility-moda}
\end{table}

\begin{table}[!t]
    \centering
    \small
    \caption{Comparison with state-of-the-art detectors on HOD3K.
    }
    \setlength{\tabcolsep}{1.5pt}
        \begin{tabular}{l|c c c|c c}
            \toprule
            \textbf{Method} & \textbf{mAP$_{50}\uparrow$} & \textbf{mAP$_{75}\uparrow$} & \textbf{mAP$\uparrow$} & \textbf{FLOPs$\downarrow$} & \textbf{Params$\downarrow$} \\
            \midrule
            D-DETR \cite{deformable-detr} & 85.0 & 52.0 & 48.8 & 109.6G & 40.1M \\
            CO-DETR \cite{zong2023detrs} & 77.9 & 57.3 & 51.7 & 200.7G & 64.5M \\
            RetinaNet \cite{retinanet} & 91.2 & 59.4 & 53.3 & 205.7G & 36.2M \\
            YOLOv5 \cite{yolov5} & 88.1 & 59.8 & 54.4 & \textbf{48.3G} & \textbf{20.9M} \\
            DINO \cite{dino} & 89.1 & 61.6 & 56.3 & 114.7G & 47.6M \\
            Fovea \cite{foveabox} & 92.2 & 62.2 & 56.8 & 123.9G & 38.0M \\
            Faster RCNN \cite{fasterrcnn} & 89.4 & 62.7 & 56.9 & 206.7G & 41.1M \\
            ATSS \cite{ATSS} & 91.2 & 63.0 & 57.4 & 110.6G & 32.2M \\
            TOOD \cite{tood} & 90.3 & 65.6 & 57.8 & 108.5G & 32.1M\\
            Meth. Mapper \cite{kumar2023methanemapper} & 88.7 & 63.8 & 57.9 & 267.6G & 80.3M \\
            RepPoints \cite{reppoints} & 92.0 & 65.5 & 58.4 & 104.5G & 36.9M\\
            PAA \cite{paa} & 91.2 & 65.0 & 58.4 & 110.6G & 32.2M \\
            VFNet \cite{varifocalnet} & 92.3 & 66.5 & 59.0 & 104.5G & 32.8M \\
            C-RCNN \cite{Cascade} & 90.9 & 66.9 & 59.6 & 149.5G & 69.2M \\
            S2ADet \cite{he2023object}& 93.4 & - & 59.8 & 169.2G & 48.6M \\
            OSSDet \cite{han2025moda} & 93.4 & 68.8 & 60.9 & 136.2G & 37.7M \\
            \midrule
            \rowcolor{tablepurple}
SGLDet (Ours) & \textbf{94.0} & \textbf{69.4} & \textbf{61.3} & 127.8G & 37.2M \\
            \bottomrule
        \end{tabular}
    \label{tab3}
\end{table}

\begin{figure}[!t]
    \centering
    \includegraphics[width=\linewidth]{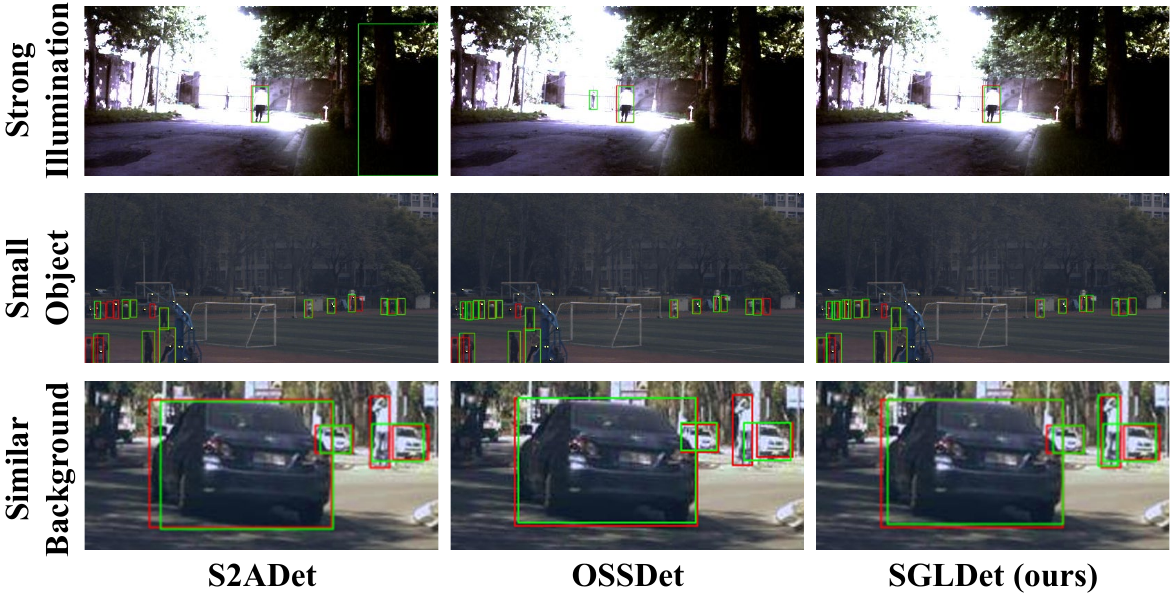}
    \vspace{-5mm}
    \caption{Detection results on HOD3K. \textcolor{red}{Red} boxes indicate the ground truth, while \textcolor{updategreen}{green} boxes denote the predictions.}
    \vspace{-2mm}
    \label{fig:hod3k_result_vis}
\end{figure}
\begin{figure}[t]
    \centering
    \includegraphics[width=\linewidth]{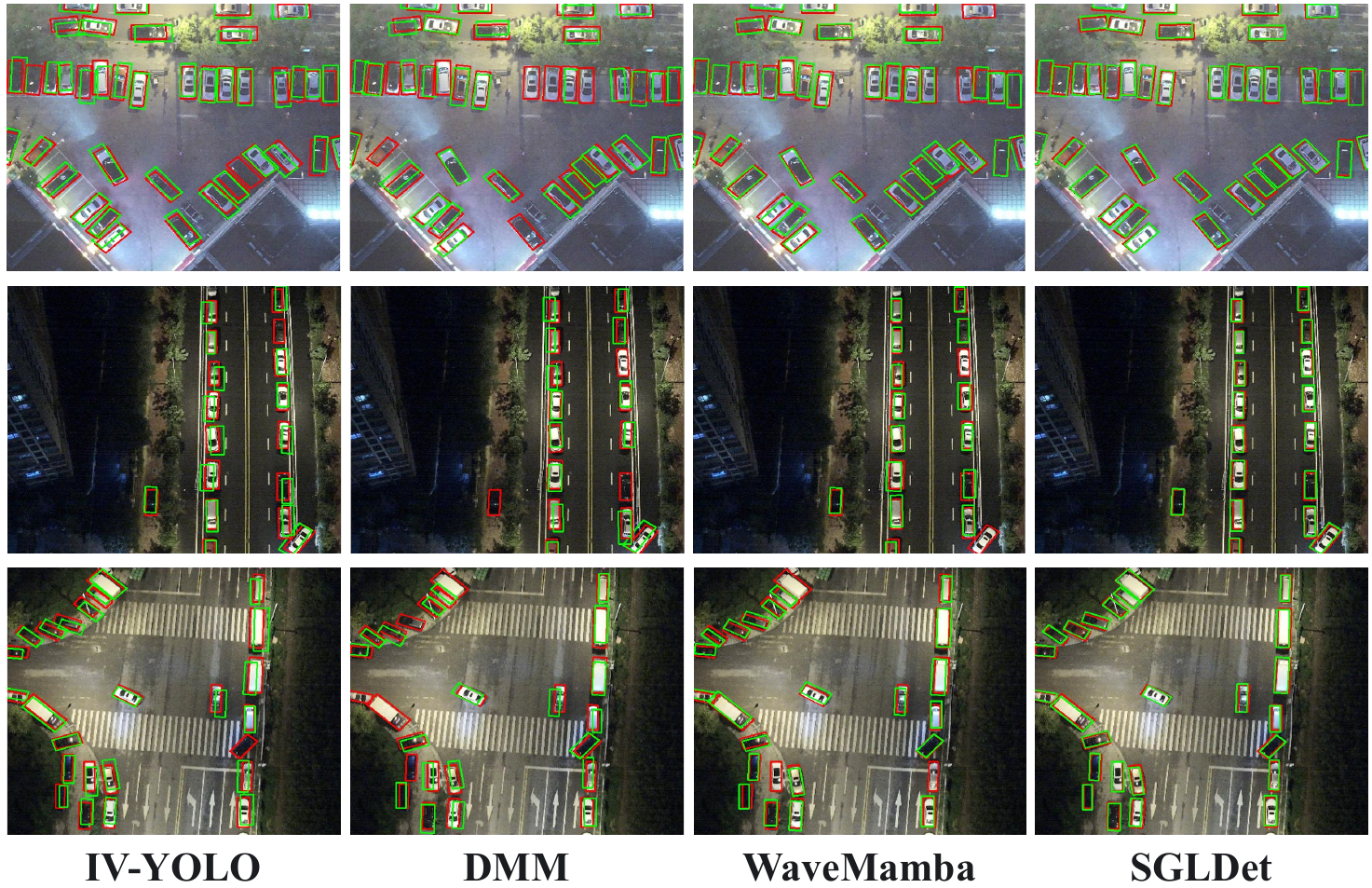}
    \caption{Detection results on DroneVehicle. \textcolor{red}{Red} boxes indicate the ground truth, while \textcolor{updategreen}{green} boxes denote the predictions.}
    \label{fig:dronevehicle_result_vis}
\end{figure}

\subsection{Results on the HOD3K Dataset}
\noindent\textbf{Data and Setups.}
The HOD3K dataset~\cite{he2023object} is a multi-band spectral object detection benchmark collected from urban roads, campuses, and residential areas. It contains 3,242 images with 16 spectral bands from 470 to 620 nm and a spatial resolution of $512 \times 256$. The dataset provides 15,149 annotated instances from three categories (people, cars, and bikes) and is split into 2,308 training, 219 validation, and 715 testing images. All 16 spectral bands were used as input, and training was conducted for 12 epochs with a total batch size of 8 at an input resolution of $1024 \times 512$. Other optimization settings followed those used on MODA, while axis-aligned IoU was adopted for evaluation.

\noindent\textbf{Quantitative Results.}
As shown in \cref{tab3}, SGLDet achieves state-of-the-art performance across all metrics, outperforming the second-best method by 0.6\%, 0.6\%, and 0.4\% in mAP$_{50}$, mAP$_{75}$, and mAP, respectively. S2ADet obtains competitive accuracy via a dual-stream spectral–spatial decoupling design, but incurs additional preprocessing and computational overhead. OSSDet directly processes MSI data and leverages object-aware guidance to learn target-relevant spectral–spatial features, improving efficiency while maintaining strong performance. In contrast, SGLDet conducts unified global and local dynamic modeling under joint spectral–spatial guidance. With cost comparable to OSSDet, it achieves higher accuracy, demonstrating more effective spectral–spatial utilization and a better efficiency–performance trade-off. Small, scale-varying aerial targets~\cite{jia2025airborne,li2023automated} and low-light conditions~\cite{jalali2022vevid} complicate detection. On HOD3K, with larger relative targets and more stable backgrounds~\cite{han2025moda}, SGLDet's improvement is less pronounced than on MODA.

\noindent\textbf{Qualitative Results.}
As shown in~\cref{fig:hod3k_result_vis}, existing methods often fail to reliably detect small objects embedded in visually similar backgrounds. This limitation stems from insufficient exploitation of spectral–spatial cues under challenging conditions. In contrast, SGLDet enhances target discrimination through unified global–local dynamic modeling guided by spectral–spatial cues, enabling more effective separation of foreground objects from background clutter and producing results that align closely with ground truth.

\subsection{Results on the DroneVehicle Dataset}
\noindent\textbf{Data and Setups.}
The DroneVehicle dataset~\cite{dronevehicle} is a widely used RGB-infrared aerial benchmark for drone-based vehicle detection. It contains 28,439 paired visible--infrared images and 953,087 oriented annotations over 5 vehicle categories, with 17,990 training pairs, 1,469 validation pairs, and 8,980 testing pairs under the official split. Instead of using two-stream fusion, the original SGLDet concatenates each RGB-infrared pair along the channel dimension as a unified four-channel spectral input within the same single-stream framework. This design exploits complementary cues from both modalities through a shared encoder without introducing a separate infrared branch. The input image size was set to $640 \times 512$, and the model was trained for 12 epochs with a total batch size of 16. Other optimization settings followed those used on MODA.

\noindent\textbf{Quantitative Results.}
\Cref{tab:dronevehicle} shows that the original four-channel model achieves 81.5\% mAP$_{50}$ and 61.8\% mAP, exceeding MS2Fusion by 1.3\% and 1.4\%, respectively. The dual stem applies separate $7\times7$ convolutions to RGB and infrared, concatenates their outputs, and projects them to 64 channels with a $1\times1$ convolution before the shared backbone. The original model leads on bus; the dual stem leads on truck and freight car, while IV-YOLO and DMM lead on car and van. Early fusion improves all five class scores over both single-modal settings. Against RGB-only, it gains 2.1\% in mAP$_{50}$ and 3.0\% in mAP. Against IR-only, it gains 1.1\% in mAP$_{50}$ and 1.5\% in mAP. The dual stem reaches 81.7\% mAP$_{50}$ and 61.9\% mAP, adding 0.2\% and 0.1\%, respectively. Prior RGB--infrared detection comparisons also found that fusing modality features after extraction performed slightly better than input fusion~\mbox{\cite{woo2024mtmmc,cao2021attention}}.

\Needspace{10\baselineskip}
\noindent\textbf{Qualitative Results.}
As illustrated in~\cref{fig:dronevehicle_result_vis}, representative RGB-infrared detectors still exhibit missed detections and false responses in densely distributed traffic scenes, especially when vehicles are small, heavily overlapped, or affected by illumination variation across modalities. In contrast, SGLDet produces more complete and accurate predictions under these challenging conditions. This qualitative advantage is consistent with the proposed global-local dynamic modeling strategy, which couples spectrally guided global reasoning with scene-aware local refinement to better preserve target completeness while suppressing cross-modal background interference.

\begin{table}[!t]
    \centering
    \small
    \caption{Performance comparisons on DroneVehicle.}
    \setlength{\tabcolsep}{1.2pt}
    \begin{tabular}{l|c c c c c|c c}
        \toprule
\textbf{Method} & \textbf{Car} & \textbf{Bus} & \textbf{Truck} & \textbf{Fre.} & \textbf{Van} & \textbf{mAP$_{50}\uparrow$} & \textbf{mAP$\uparrow$} \\
        \midrule
        C$^2$Former \cite{yuan2024c2former} & 90.2 & 89.8 & 68.3 & 64.4 & 58.5 & 74.2 & 47.3 \\
        YOLOv8l-RGB \cite{yolov8} & 92.5 & 91.7 & 68.8 & 47.8 & 50.2 & 70.2 & 48.6 \\
        YOLOv8l-IR \cite{yolov8} & 93.4 & 91.9 & 69.3 & 53.7 & 51.1 & 71.9 & 49.1 \\
        SLBAF-Net \cite{cheng2023slbaf} & 90.2 & 89.9 & 76.0 & 68.2 & 59.9 & 76.8 & 49.5 \\
        CSOM-ODAF \cite{chen2024weakly} & 90.1 & 89.8 & 75.6 & 68.2 & 61.8 & 77.1 & 50.1 \\
        Multi. DINO \cite{sun2024freqgate} & 89.5 & 88.8 & 75.4 & 54.3 & 54.3 & 72.5 & 50.3 \\
        CMA \cite{jiang2024m2fnet} & 95.8 & 93.1 & 75.9 & 59.8 & 59.4 & 76.8 & 50.4 \\
        CRSIOD \cite{wang2024cross} & 95.6 & 92.2 & 71.7 & 50.5 & 55.8 & 73.2 & 50.8 \\
        GLFNet \cite{kang2024glfnet} & 90.3 & 88.0 & 72.7 & 53.6 & 52.6 & 71.4 & 54.8 \\
        Dual-YOLO \cite{bao2023dual} & 95.9 & 91.6 & 69.7 & 55.9 & 46.6 & 71.9 & 55.2 \\
        DMM \cite{zhou2024dmm} & 90.4 & 88.7 & 77.8 & 63.0 & \textbf{66.0} & 77.2 & 55.8 \\
        IV-YOLO \cite{tian2024ivyolo} & \textbf{97.2} & 94.3 & 65.4 & 63.1 & 53.0 & 74.6 & 56.8 \\
        WaveMamba \cite{zhu2025wavemamba} & 95.0 & 90.6 & 80.4 & 68.5 & 64.5 & 79.8 & 60.5 \\
        O-RCNN \cite{xie2021oriented} & 90.8 & 90.0 & 66.8 & 60.2 & 50.2 & 71.6 & 44.3 \\
        S$^2$ANet \cite{han2021align} & 90.6 & 89.5 & 66.2 & 61.8 & 52.9 & 72.2 & 43.8 \\
        OSSDet~\cite{han2025moda} & 94.0 & 89.5 & 78.0 & 67.0 & 63.0 & 78.3 & 58.5 \\
        ICAFusion \cite{shen2024icafusion} & 95.6 & 92.3 & 77.8 & 64.5 & 60.3 & 78.1 & 58.6 \\
        SFFR \cite{zuo2025sffr} & 95.8 & 93.1 & 79.2 & 66.1 & 62.3 & 79.3 & 59.4 \\
        MS2Fusion \cite{shen2026ms2fusion} & 96.1 & 93.7 & 80.5 & 67.8 & 62.9 & 80.2 & 60.4 \\
        IRDFusion \cite{shen2026irdfusion} & 95.7 & 92.2 & 79.4 & 66.3 & 64.1 & 79.5 & 59.7 \\
        \midrule
        SGLDet-RGB & 95.8 & 93.9 & 78.9 & 65.5 & 62.9 & 79.4 & 58.8 \\
        SGLDet-IR & 95.6 & 93.8 & 80.3 & 67.9 & 64.4 & 80.4 & 60.3 \\
\rowcolor{lightpurple}SGLDet & 96.5 & \textbf{94.8} & 81.2 & 69.7 & 65.5 & 81.5 & 61.8 \\
        SGLDet-DualStem & 96.6 & 94.7 & \textbf{81.5} & \textbf{69.9} & 65.8 & \textbf{81.7} & \textbf{61.9} \\
        \bottomrule
    \end{tabular}
    \label{tab:dronevehicle}
\end{table}

%% file: sec/table_moda_challenging_subsets_main.tex
\begin{table}[!h]
\centering
\footnotesize
\caption{Challenging-subset comparison on MODA.}
\setlength{\tabcolsep}{1.0pt}
\renewcommand{\arraystretch}{1.06}
\begin{tabular}{c|ccc|ccc}
\toprule
\multirow{2}{*}{\textbf{Subset}} & \multicolumn{3}{c|}{\textbf{OSSDet}} & \multicolumn{3}{c}{\textbf{SGLDet (Ours)}} \\
& \textbf{mAP$_{50}\uparrow$} & \textbf{mAP$_{75}\uparrow$} & \textbf{mAP$\uparrow$} & \textbf{mAP$_{50}\uparrow$} & \textbf{mAP$_{75}\uparrow$} & \textbf{mAP$\uparrow$} \\
\midrule
Full test       & 69.0 & 45.9 & 42.7 & \textbf{71.8} & \textbf{47.0} & \textbf{44.3} \\
\midrule
Low illum.      & 61.4 & 38.2 & 35.5 & \textbf{65.1} & \textbf{40.8} & \textbf{38.0} \\
Small objects   & 56.2 & 32.5 & 30.8 & \textbf{59.0} & \textbf{34.8} & \textbf{33.0} \\
High density    & 63.8 & 41.0 & 38.5 & \textbf{67.0} & \textbf{43.5} & \textbf{40.8} \\
\bottomrule
\end{tabular}
\label{tab:moda-challenging-subsets-main}
\end{table}

%% file: sec/table_inference_efficiency_compact.tex
\begin{table}[H]
\centering
\small
\caption{Measured inference efficiency and floating-point operation count on MODA.}
\setlength{\tabcolsep}{2.2pt}
\renewcommand{\arraystretch}{1.06}
\begin{tabular}{lccccc}
\toprule
\textbf{Method} & \textbf{mAP$_{50}\uparrow$} & \textbf{mAP$\uparrow$} & \textbf{FLOPs$\downarrow$} & \textbf{ms/img$\downarrow$} & \textbf{FPS$\uparrow$} \\
\midrule
O-RCNN~\cite{xie2021oriented}             & 65.5 & 40.9 & 244.7G & $51\!\pm\!0.7$ & 19.6 \\
Rot. Retina~\cite{retinanet}              & 53.4 & 34.1 & 233.5G & $42\!\pm\!0.6$ & 23.8 \\
Ori. RepPoints~\cite{oriented-reppoints}  & 66.5 & 40.9 & 213.6G & $56\!\pm\!0.8$ & 17.9 \\
Ori-DETR~\cite{zhao2024pointaxis}         & 67.8 & 41.9 & 492.0G & $76\!\pm\!1.5$ & 13.2 \\
S2ADet~\cite{he2023object}                & 63.5 & 38.9 & 406.0G & $112\!\pm\!1.8$ & 8.9 \\
OSSDet~\cite{han2025moda}                 & 69.0 & 42.7 & 263.1G & $53\!\pm\!0.8$ & 18.9 \\
\midrule
\textbf{SGLDet (Ours)} & \textbf{71.8} & \textbf{44.3} & 241.6G & $54\!\pm\!0.8$ & 18.5 \\
\bottomrule
\end{tabular}
\label{tab:inference-efficiency}
\end{table}

%% file: sec/table_sgdm_affinity.tex
\newcommand{\sgdmaffinitytabular}{%
\begin{tabular}{l|ccc}
\toprule
Affinity function & mAP$_{50}\uparrow$ & mAP$_{75}\uparrow$ & mAP$\uparrow$ \\
\midrule
Cosine similarity & 71.1 & 46.3 & 43.5 \\
Spectral Angle Mapper & 70.7 & 46.0 & 43.1 \\
\rowcolor{lightpurple}Euclidean affinity & \textbf{71.8} & \textbf{47.0} & \textbf{44.3} \\
\bottomrule
\end{tabular}%
}

%% file: sec/5_conclusion.tex
\section{Conclusion}
\label{sec:conclu}
We propose SGLDet, a spectral-aware global-to-local dynamic modeling framework for multispectral aerial object detection. To address unreliable recognition caused by severe foreground-background imbalance and inaccurate localization caused by substantial variations in target scale, orientation, and structure, SGLDet integrates three complementary modules. SGDM captures long-range spectral dependencies through similarity-guided sparse retrieval to improve semantic consistency among spectrally similar objects. MGCI consolidates multi-scale evidence into a compact scene memory for holistic scene understanding. SICA further performs dynamic feature modulation via input-adaptive expert aggregation and spectral-spatial calibration to handle pronounced local variations. Extensive experiments on MODA, HOD3K, and DroneVehicle demonstrate the effectiveness and generalization ability of SGLDet across diverse multispectral detection scenarios. Future work will explore extending SGLDet's global-to-local spectral-spatial modeling to other multispectral aerial perception tasks.